\documentclass{kyrielei}
\usepackage{amsmath,amssymb}
\usepackage{algorithm2e}
\usepackage{enumitem}
\usepackage{wrapfig}
\usepackage{hyperref}
\usepackage{url}
\usepackage{booktabs}
\usepackage{graphicx}
\usepackage[table]{xcolor}
\usepackage{multirow}
\usepackage{wrapfig}
\usepackage{graphicx}
\usepackage{subcaption}
\usepackage{xspace}
\usepackage{comment}
\usepackage{adjustbox}
\usepackage{siunitx}

\title{Just MLPs: Efficient Visual State Reconstruction for Multimodal Language Models}
\logo{\includegraphics[height=12mm]{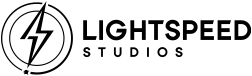}}
\secondarylogos{%
  \includegraphics[height=12mm]{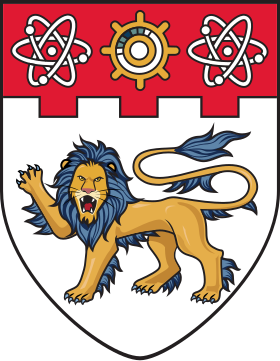}%
}

\newcommand{\method}{\ensuremath{\delta\text{-Vision}}\xspace}
\definecolor{myblue}{RGB}{224,242,255}
\definecolor{qwenpurple}{HTML}{EEE7FF}  
\definecolor{llavaorange}{HTML}{FFE8CC} 

\author[1]{Jingdi Lei}
\author[2]{Junxian Li}
\author[3]{Di Zhang}
\author[4]{Zhanqiu Zhang}
\author[5]{Yiwen Guo}
\author[1]{Soujanya Poria}
\affiliation[1]{Nanyang Technological University}
\affiliation[2]{Shanghai Jiao Tong University}
\affiliation[3]{Fudan University}
\affiliation[4]{LIGHTSPEED}
\affiliation[5]{Independent Researcher}

\metadata[{\faGithub\ Github}]{
  \href{https://github.com/declare-lab/delta-Vision}{DeCLaRe Lab}
}
\metadata[{\faEnvelope\ Correspondence}]{
  \href{jingdi001@e.ntu.edu.sg}{Jingdi Lei}, \href{zqzhang27@gmail.com}{Zhanqiu Zhang}, \href{guoyiwen89@gmail.com}{Yiwen Guo}, \href{soujanya.poria@ntu.edu.sg}{Soujanya Poria},
}

\abstract{
\vspace{0.8em}
Long visual token sequences often account for a substantial fraction of the computational overhead in multimodal large language models~(MLLMs). Existing approaches reduce this cost by pruning redundant visual tokens, but permanently discard visual evidence that may become useful in subsequent layers. We instead ask whether all visual tokens can be preserved while reducing the cost of repeatedly evolving the representations through the Transformer. To answer this question, we perform low-rank interventions on visual-to-text information flow. We find that, after visual-to-text attention is blocked, restoring only a few directions recovers most of the lost accuracy, suggesting the relevant visual influence is concentrated in a low-dimensional subspace. We further observe strong predictability in layer-specific visual states: lightweight MLPs approximate them with high cosine similarity and low reconstruction error. Motivated by these findings, we propose \method, which replaces repeated Transformer evolution of visual tokens with lightweight low-rank adapters that construct layer-wise visual memories while preserving all visual tokens for text retrieval. Across image and video benchmarks, \method achieves higher accuracy than visual token pruning baselines at comparable or lower computation, while delivering competitive inference efficiency without discarding visual tokens.
}

\date{\today}

\begin{document}
\maketitle

\section{Introduction}
\label{Intro}
Multimodal large language models~(MLLMs) have extended large language models~(LLMs) beyond textual reasoning by endowing them with visual perception and understanding~\citep{liu2023visual,hurst2024gpt,bai2025qwen3,team2025kimi,hong2026glm}, and have demonstrated remarkable capabilities across a diverse range of multimodal tasks, including image captioning~\citep{alayrac2022flamingo}, visual question answering~(VQA), video understanding~\citep{bai2025qwen3}, and multimodal reasoning~\citep{yue2024mmmu}. However, such impressive performance is accompanied by substantial computational overhead, particularly for high-resolution images and videos which are represented by long sequences of visual tokens~\citep{yang2025visionzip, shang2025llava}. This computational cost is further amplified because these visual tokens participate in computations at every single Transformer layer of the inner LLM. As a result, efficiently handling long visual token sequences has become an important challenge for scalable multimodal inference. 

Existing approaches~\citep{yang2025visionzip, wen2025stop, wen2026efficient} have explored reducing this cost by pruning redundant visual tokens, thereby shortening the visual sequence processed by the language model. While effective, such removal is irreversible, once a visual token is discarded, the corresponding visual evidence can no longer be accessed by subsequent layers~\citep{chen2026one, yang2026reroute, qian2026swiftvlm}. Rather than whether asking visual tokens can be safetly discarded, we explore a complementary direction: preserving all visual tokens while reducing their processing cost within the LLM. Specifically, do visual tokens require full computation at every Transformer layer, or can the necessary visual information be integrated into the text stream more efficiently? 

\begin{figure*}[t]
    \centering
    \begin{subfigure}[htbp]{0.48\textwidth}
        \centering
        \includegraphics[
            width=\linewidth,
            trim=5 5 5 5,
            clip
        ]{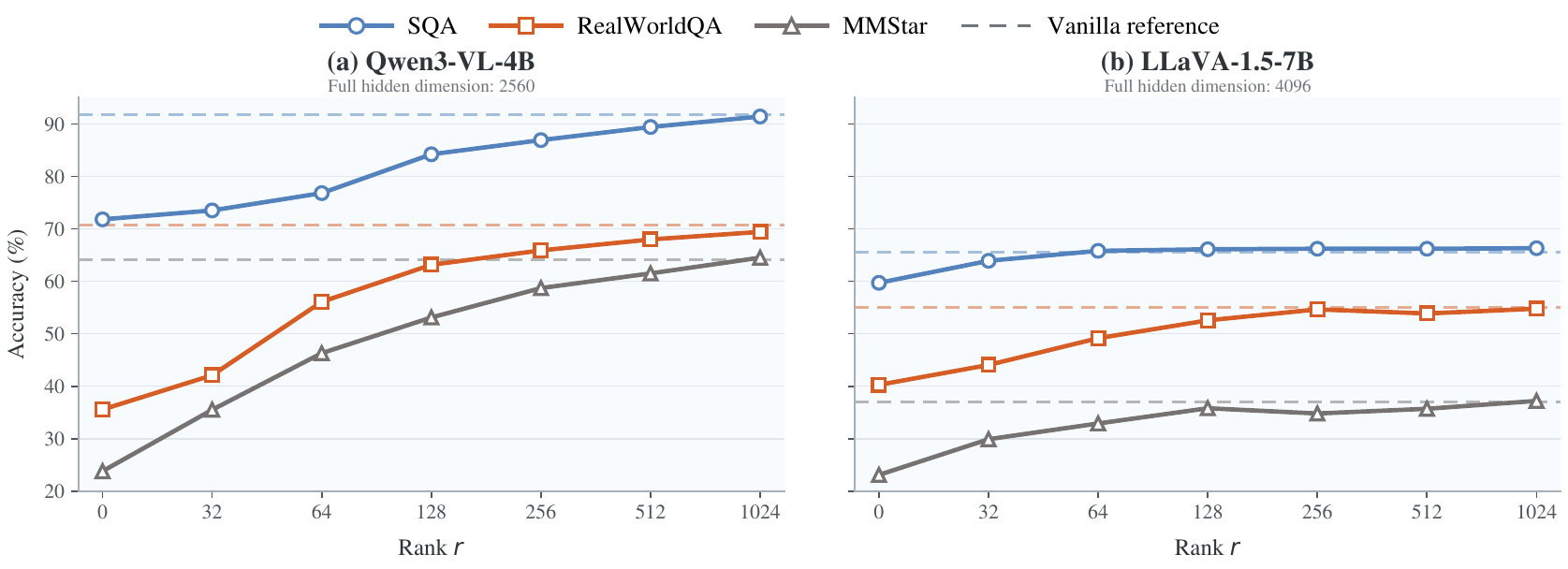}
        \caption{
        Accuracy recovery under low-rank approximation.
        }
        \label{fig:rank_accuracy_recovery}
    \end{subfigure} 
    \hfill
    \begin{subfigure}[htbp]{0.48\textwidth}
        \centering
        \includegraphics[
            width=\linewidth,
            trim=5 5 5 5,
            clip
        ]{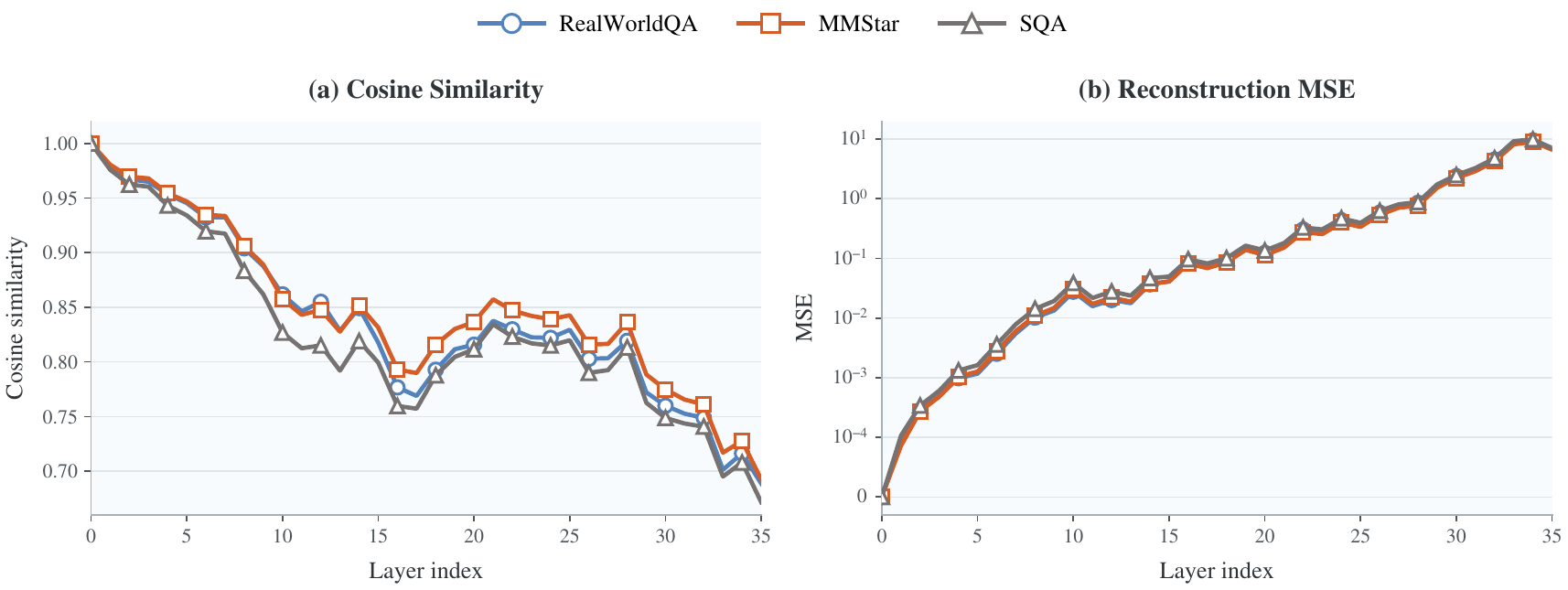}
        \caption{
        Layer-wise predictability of visual representations.
        }
        \label{fig:layerwise_predictability}
    \end{subfigure}
    \caption{
    Analysis of the dimensional structure of visual representations.
    (a) Downstream accuracy under different retained ranks of visual representation.
    (b) Layer-wise reconstruction quality measured by cosine similarity and reconstruction MSE.
    }
    \label{fig:low_dimensional_visual_analysis}
\end{figure*}

To investigate this possibility, we first examine effective dimensionality required for visual information to influence the text stream. Keeping all visual tokens and their positions unchanged, we project each visual hidden state onto a low-rank subspace and reconstruct it before subsequent computation~(Details are provided in Appendix~\ref{appendix:hidden_state_compression}). As shown in Figure~\ref{fig:rank_accuracy_recovery}, much of the original model performance can be retained at ranks substantially smaller than the native hidden dimension, particularly for LLaVA-1.5-7B~\citep{liu2023visual}. This observation suggests that visual redundancy exists not only across tokens, but also in the computation used to update their representations across layers. More importantly, these layer-specific visual states also exhibit strong predictability. We train a separate lightweight MLP with a $d\!\rightarrow\!d\!\rightarrow\!d$ architecture for each layer to predict its visual states from the initial visual embeddings. As shown in Figure~\ref{fig:layerwise_predictability}, these MLPs approximate the Transformer-evolved states with high cosine similarity and low reconstruction error. These observations suggest that preserving useful visual information may not require repeatedly computing full-dimensional visual states through every Transformer layer.

Motivated by these observations, we develop \method. Rather than propagating visual tokens through full self-attention and feed-forward computation, \method~directly constructs the visual memory required at each layer using a low-rank adapter, while keeping the language-model backbone frozen. We instantiate this idea with an embedding adapter that predicts each layer’s visual memory directly from the initial embeddings and a recurrent adapter that progressively updates the predicted memory across layers. At each layer, the predicted memory is mapped through the original frozen key and value projections. Consequently, text tokens retain access to every visual token, while visual queries, visual attention outputs, and visual feed-forward computation are removed. In this way, \method~changes how visual states are constructed rather than which visual evidence remains accessible, providing an alternative efficiency axis to visual token pruning. 

Extensive experiments across different MLLM backbones and a broad range of image, multi-image, and video benchmarks demonstrate that \method~achieves a strong accuracy-efficiency trade-off. On Qwen3-VL-4B, \method~reaches an average score of 74.4 under a computational budget comparable to pruning methods that discard 95\% of visual tokens, outperforming the strongest baseline by 9.1 points and even matching the best pruning result obtained while retaining 20\% of the tokens. The effectiveness of visual-memory prediction further generalizes across different scales and multi-image benchmarks. 
On Video-MME, \method~requires only 17.90\% of the FLOPs of the uncompressed model and achieves $1.30\times$ total speed and $1.50\times$ prefill speedup, while maintaining a prefill efficiency comparable to token-pruning baselines despite preserving all visual tokens.

Our contributions can be summarized as follows:
\begin{itemize}
    \item We provide empirical evidence that preserving visual information does not necessarily require full layer-wise Transformer computation. By keeping all visual tokens, we show that layer-wise visual states are highly compressible in the hidden-channel dimension and exhibit strong predictability across layers.
    \item We propose \method, a lightweight visual-memory prediction framework that replaces repeated visual-token Transformer updates with low-rank layer-wise adapters while preserving all visual tokens for text retrieval
    \item We demonstrate the effectiveness and generality of \method~across different MLLM backbones and diverse image, multi-image, and video benchmarks, achieving stronger accuracy--efficiency trade-offs than visual token pruning methods under comparable computational budgets.
\end{itemize}

\section{Preliminaries}
\label{Preliminaries}
Consider a multimodal large language model that receives a visual input $\mathbf{I}$ and a textual prompt $x_{1:T}$. A vision encoder followed by a multimodal projector converts the visual input into $N_v$ visual embeddings:
\begin{equation}
    E = [e_1, ..., e_{N_v}] \in \mathbb{R}^{N_v \times d},
\end{equation}

where $d$ is the hidden dimension of the language model. The visual embeddings are concatenated with the text embeddings and processed by an $L$-layer Transformer. Let
\begin{equation} 
H_l = [V_l; T_l] \in \mathbb{R}^{(N_v + N_t) \times d}, 
\qquad V_l \in \mathbb{R}^{N_v \times d}, 
\qquad T_l \in \mathbb{R}^{N_t \times d}, 
\end{equation}
denote the hidden states entering layer $l$, where $V_l$ and $T_l$ correspond to the visual and textual states. At the input layer, $V_0 = E$. Visual tokens play two roles in a MLLM. First, they serve as visual context for the language stream: textual queries retrieve visual information through the keys and values associated with visual tokens. Second, visual tokens are themselves active Transformer states. At every layer, they generate their own queries, receive attention outputs, pass through the feed-forward network, and are consequently transformed into new layer-specific visual states.

We refer to the sequence:
\begin{equation}
    E = V_0 \rightarrow V_1 \rightarrow \cdot\cdot\cdot \rightarrow V_L,
\end{equation}
produced by the Transformer as the layer-wise evolution of visual tokens. Importantly, the representation exposed to the text stream is layer dependent rather than directly from the initial visual embeddings $E$. This design naturally allows visual representations to evolve jointly with the language stream, but it also requires all $N_v$ visual tokens to undergo attention and feed-forward computation at every layer. For high-resolution images or videos with long visual sequences, this repeated computation becomes a substantial component of multimodal inference cost. Our goal is not to reduce $N_v$, but to reduce the computation to obtain the layer-specific visual representation.

\section{\method}
\label{Method}
\begin{figure*}[t]
    \centering
    \includegraphics[width=0.9\textwidth]{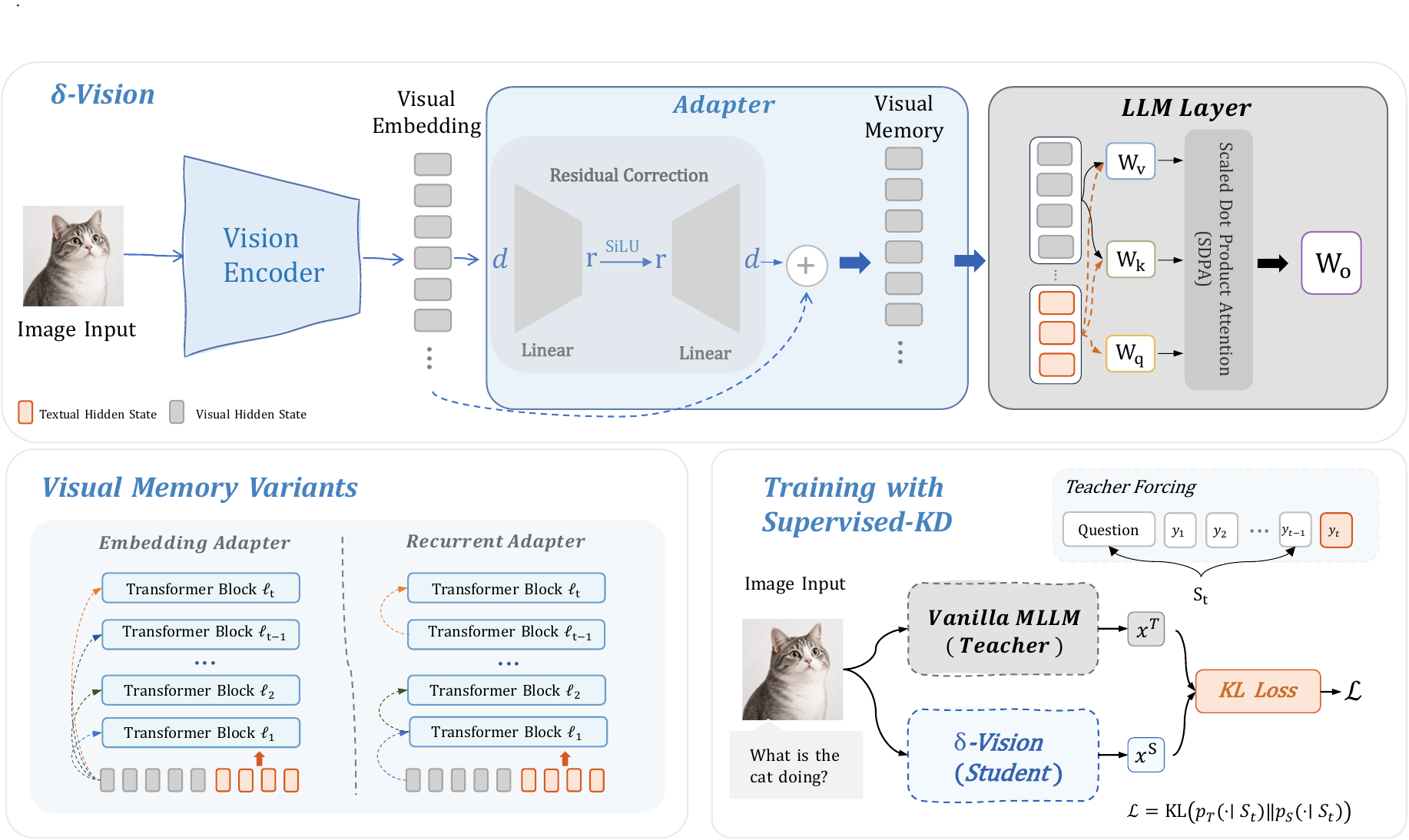}
    \caption{
    \textbf{Overview of \method.}
    Top: \method~replaces the original visual hidden-state evolution
    with lightweight low-rank adapters that constructs layer-wise visual
    representations.
    Bottom left: we consider two visual-memory variants:
    \emph{Embedding Adapter}, which predicts visual states from the initial
    visual embeddings, and \emph{Recurrent Adapter}, which propagates a
    compact visual state across layers.
    Bottom right: \method~is trained with Supervised-KD, where the frozen
    vanilla MLLM serves as the teacher and provides token-level distributional
    supervision under teacher forcing.
    }
    \label{fig:delta_vision_overview}
\end{figure*}

\method~replaces the original visual-token update pathway with lightweight layer-wise visual memory prediction, while keeping the textual part unchanged. Figure~\ref{fig:delta_vision_overview} provides an overview of this design. At layer $l$, a low-rank adapter constructs a visual memory $M_l$, which is then projected by the key--value modules and accessed by text queries. Visual queries, visual attention outputs, and visual feed-forward updates are omitted. We develop two variants for constructing $M_l$: embedding adapter that predicts the visual memory from the initial visual embeddings, and recurrent adapter that updates it from the preceding visual memory.

\subsection{Layer-Wise Visual Memory Prediction}

We parameterize the layer-wise visual memory as a full-dimensional residual correction. Given an input visual state $X \in \mathbb{R}^{N_v \times d}$, the adapter at layer $l$ is defined as
\begin{equation}
    A_l(X) = X + \Delta_l(X), \quad \Delta_l(X) = \phi(XD_l)U_l,
\end{equation}
where $D_l \in \mathbb{R}^{d\times r}$ and $U_l \in \mathbb{R}^{r\times d}$ are the down- and up-projection matrices, $r \ll d$ is the bottleneck dimension, and $\phi(\cdot)$ denotes the SiLU activation function. We initialize $U_l$ to zero, so that $A_l$ starts as an identity mapping. 

The key property of this parameterization is that it constrains how the visual memory is updated rather than the memory itself. For every visual token, the residual update $\Delta_l(X)$ lies in the subspace spanned by $U_l$, whose dimension is at most $r$. Equivalently, $rank\left(\Delta_l(X)\right) \le r$. $A_l(X)$ remains a full $d$-dimensional representation through the residual connections. \method assumes that the layer-specific displacement required to adapt visual information for a given Transformer can be represented within a low-dimensional hidden-channel subspace. 

We instantiate this low-dimensional update in two forms:

\paragraph{Embedding Adapter.} The embedding variant predicts the memory of each layer directly from the initial visual embeddings,
\begin{equation}
    M_l = A_l(E) = E + \Delta_l(E).
\end{equation}
Every layer learns its own low-dimensional displacement from the same initial visual representation. The resulting memories are therefore conditionally independent across layers given $E$, allowing all $M_l$ to be constructed without explicitly propagating visual states through preceding Transformer layers. This formulation represents our hypothesis: useful layer-specific visual states can be obtained as lightweight corrections to the initial visual evidence.

\paragraph{Recurrent Adapter.} The recurrent variant instead constructs a trajectory of visual memories through successive low-dimensional corrections,
\begin{equation}
    M_0 = E, M_l = A_l(M_{l-1}) = M_{l-1} + \Delta_l(M_{l-1}), l = 1,...,L.
\end{equation}
Unlike the embedding variant, this formulation preserves an explicit layer-to-layer state. However, each transition is restricted to a low-dimensional displacement rather than a full Transformer update. The recurrent construction can therefore be viewed as a lightweight approximation to visual-state evolution, in which the memory trajectory is formed by accumulating a sequence of compact layer-specific corrections.

\subsection{Visual Memory as Context}
Having constructed the layer-specific visual memory $M_l$, we use it as read-only key--value context for the textual stream. Specifically, the predicted memory is processed by the original frozen layer normalization and key--value projections,
\begin{equation}
K_{l}^{v}
=
W_{K,l}\operatorname{LN}_{l}(M_{l}),
\qquad
V_{l}^{v}
=
W_{V,l}\operatorname{LN}_{l}(M_{l}).
\end{equation}
Let
$H_{l} \in \mathbb{R}^{N_t \times d}$
denote the textual hidden states entering layer $l$.
Text tokens follow the original Transformer computation and produce
\begin{equation}
Q_{l}^{t}
=
W_{Q,l}\operatorname{LN}_{l}(H_{l}),
\qquad
K_{l}^{t}
=
W_{K,l}\operatorname{LN}_{l}(H_{l}),
\qquad
V_{l}^{t}
=
W_{V,l}\operatorname{LN}_{l}(H_{l}).
\end{equation}
The textual queries then attend jointly to the predicted visual memory
and the textual context,
\begin{equation}
O_{l}^{t}
=
\operatorname{Attn}
\left(
Q_{l}^{t},
[K_{l}^{v}; K_{l}^{t}],
[V_{l}^{v}; V_{l}^{t}]
\right),
\end{equation}
where the original causal attention structure is preserved.
The resulting attention output contains only $N_t$ query positions
and is subsequently processed by the original output projection,
residual connection, and feed-forward network of the frozen language model.

This formulation separates visual-state construction from visual information retrieval. In a conventional MLLM, visual tokens simultaneously provide keys and values to the text stream and act as active Transformer states. In \method, only the former role is retained. The predicted memory contributes $K_{l}^{v}$ and $V_{l}^{v}$ at every layer, so textual queries can still retrieve information from every visual token, but no $Q_{l}^{v}$, visual attention output, or visual feed-forward update is computed.

Importantly, $M_l$ is not updated by the textual attention computation. Once consumed as key--value context at layer $l$, it is discarded, and the memory for the next layer $M_{l+1}$ is obtained directly from the corresponding embedding or recurrent adapter. Thus, visual memories form an external layer-wise context sequence rather than a set of hidden states propagated through the Transformer backbone. This allows \method to preserve access to all visual tokens simultaneously.

\subsection{Training Objective}

We optimize only the visual-memory adapters while keeping the vision encoder, multimodal projector, and language-model backbone frozen. The original MLLM serves as the teacher, while \method acts as the student. Inspired by \citet{agarwal2024policy}, we use Supervised-KD as our training objective. Specifically, for each training example, both models are conditioned on the same multimodal input and the same ground-truth answer prefix. At answer position $t$, teacher and student therefore predict the next token $y_t$ conditioned on $\left(I, x_{1:T}, y_{<t}\right)$.

We minimize the forward KL divergence between the teacher and student predictive distributions over ground-truth answer positions,
\begin{equation}
\mathcal{L}_{\mathrm{KD}}
=
\frac{1}{|\Omega|}
\sum_{t \in \Omega}
D_{\mathrm{KL}}
\left(
p_T(\cdot \mid I, x_{1:T}, y_{<t})
\,\|\, 
p_S(\cdot \mid I, x_{1:T}, y_{<t})
\right),
\end{equation}
where $\Omega$ denotes the set of answer-token positions. Since the backbone is frozen, the supervision is absorbed entirely by the lightweight visual-memory predictors, encouraging them to provide layer-wise visual context that preserves the teacher's output behavior. An ablation against standard SFT and on-policy distillation is provided in Appendix~\ref{appendix:training_objective_ablation}, showing that Supervised-KD achieves the best average performance while avoiding the substantial rollout cost of on-policy training.

\section{Experiments}

\label{Experiments}
\subsection{Experimental Setup}

\textbf{Models and Training.} Unless otherwise specified, we use Qwen3-VL-4B-Instruct~\citep{bai2025qwen3}~(hereafter referred to as Qwen3-VL-4B) with the embedding adapter as the default \method configuration. All training configurations used in this work are provided in Appendix~\ref{appendix:additional_implementation_and_training_details}.
\begin{table}[t]
    \centering
    \caption{
    Results of visual token compression methods under 20\% and 5\% visual-token retention.
    }
    \label{tab:baseline_acc_comparison_main}
    \resizebox{\textwidth}{!}{%
    \renewcommand{\arraystretch}{1.15}
    \begin{tabular}{lcccccccccc}
        \toprule
        \textbf{Method}
        & \textbf{MMStar}
        & \textbf{RWQA}
        & \textbf{GQA}
        & \textbf{MMB}
        & \textbf{MMB-CN}
        & \textbf{MME}
        & \textbf{POPE}
        & \textbf{SQA}
        & \textbf{VQA-v2}
        & \textbf{Avg.} \\
        \midrule

        \rowcolor{gray!15}
        \multicolumn{11}{c}{\textit{Upper Bound (100\% Retention)}} \\

        Qwen3-VL-4B
        & 64.9 & 71.2 & 61.6 & 87.5 & 87.7
        & 84.7 & 89.3 & 93.3 & 80.9 & 80.1 \\

        \midrule

        \rowcolor{gray!15}
        \multicolumn{11}{c}{\textbf{Training-Free}} \\

        \rowcolor{gray!10}
        \multicolumn{11}{c}{\textit{Retain 20\% Visual Tokens}} \\

        FastV{\footnotesize\texttt{(ECCV24)}}
        & 49.5 & 55.0 & 47.3 & 79.2 & 77.8
        & 76.6 & 77.6 & 83.2 & 65.8 & 68.0 \\

        VisionZip{\footnotesize\texttt{(CVPR25)}}
        & 51.1 & 61.7 & 52.9 & 82.0 & 80.8
        & 78.6 & 82.6 & 85.2 & 70.8 & 71.7 \\

        DivPrune{\footnotesize\texttt{(CVPR25)}}
        & 53.2 & 63.9 & 55.5 & 84.2 & 83.5
        & 81.9 & 86.3 & 84.4 & 75.3 & 74.2 \\

        SparseVLM{\footnotesize\texttt{(ICML25)}}
        & 53.2 & 45.1 & 43.1 & 73.2 & 71.1
        & 66.3 & 77.8 & 76.3 & 68.1 & 63.8 \\

        DART{\footnotesize\texttt{(EMNLP25)}}
        & 51.4 & 63.9 & 53.5 & 82.8 & 82.0
        & 77.9 & 84.8 & 85.0 & 72.1 & 72.6 \\

        ZOO-Prune{\footnotesize\texttt{(CVPR26)}}
        & 49.3 & 54.1 & 50.4 & 80.5 & 79.5
        & 78.2 & 78.7 & 79.7 & 70.5 & 69.0 \\

        \rowcolor{gray!10}
        \multicolumn{11}{c}{\textit{Retain 5\% Visual Tokens}} \\

        FastV{\footnotesize\texttt{(ECCV24)}}
        & 34.8 & 43.2 & 39.9 & 59.4 & 56.6
        & 67.5 & 60.9 & 72.4 & 50.6 & 53.9 \\

        VisionZip{\footnotesize\texttt{(CVPR25)}}
        & 39.1 & 47.8 & 44.6 & 71.6 & 67.9
        & 73.5 & 66.9 & 75.2 & 59.6 & 60.7 \\

        DivPrune{\footnotesize\texttt{(CVPR25)}}
        & 42.8 & 57.1 & 45.0 & 75.9 & 76.0
        & 74.5 & 73.4 & 77.0 & 65.7 & 65.3 \\

        SparseVLM{\footnotesize\texttt{(ICML25)}}
        & 37.5 & 42.1 & 39.2 & 53.0 & 51.3
        & 57.3 & 59.5 & 72.1 & 51.0 & 51.4 \\

        DART{\footnotesize\texttt{(EMNLP25)}}
        & 42.5 & 52.0 & 43.5 & 73.6 & 73.8
        & 68.0 & 71.0 & 79.4 & 63.7 & 63.1 \\

        ZOO-Prune{\footnotesize\texttt{(CVPR26)}}
        & 37.5 & 44.8 & 42.2 & 69.9 & 69.6
        & 68.5 & 61.9 & 75.6 & 58.4 & 58.7 \\

        \midrule

        \rowcolor{gray!15}
        \multicolumn{11}{c}{
            \textbf{Training-Based} \textit{(Retain 5\% Visual Tokens)}
        } \\

        \textit{LLaVA-Mini}{\footnotesize\texttt{(ICLR25)}}
        & 40.8 & 51.1 & 47.3 & 78.2 & 78.4
        & 70.7 & 78.0 & 76.1 & 64.3 & 65.0 \\

        EPIC{\footnotesize\texttt{(NeurIPS25)}}
        & 38.4 & 58.8 & 49.2 & 77.8 & 76.2
        & 74.7 & 81.2 & 79.6 & 66.0 & 66.9 \\

        \midrule

        \rowcolor{myblue}
        \textbf{\textit{\method}(Emb. Adapter)}
        & 54.4 & 62.9 & 56.7 & 84.5 & 82.8
        & 80.1 & 87.6 & 82.3 & 78.0 & 74.4 \\

        \rowcolor{myblue}
        \textbf{\textit{\method}(Rec. Adapter)}
        & 56.9
        & 64.4
        & 58.4
        & 85.0
        & 84.6
        & 80.7
        & 87.8
        & 82.3
        & 78.5
        & \textbf{75.4} \\

        \bottomrule
    \end{tabular}%
     }
\end{table}

\textbf{Evaluations.}
We evaluate \method across single-image, multi-image, and video understanding benchmarks. For single-image evaluation, we use MMStar~\citep{chen2024we}, RealWorldQA~(RWQA)~\citep{realworldqa2024}, GQA~\citep{gqa}, MMBench~(MMB)~\citep{liu2024mmbench}, MMBench-CN~(MMB-CN)~\citep{liu2024mmbench}, MME~\citep{fu2026mme}, POPE~\citep{pope}, ScienceQA~(SQA)~\citep{lu2022learn}, VQA-v2~\citep{goyal2017making}, covering general multimodal reasoning, real-world visual understanding, compositional reasoning, scientific reasoning, and open-ended visual question answering. For more complex visual inputs, we evaluate multi-image understanding on MuirBench~\citep{wang2025muirbench} and video understanding on Video-MME~\citep{fu2025video} and MVBench~\citep{li2024mvbench}. For each benchmark, we report per-question accuracy to facilitate consistent averaging and comparison across benchmarks. The overall score is the unweighted mean of all benchmark scores.

Since \method primarily reduces the cost of processing visual tokens, we compare \method against several training-free visual token pruning methods, including FastV~\citep{chen2024image}, VisionZip~\citep{yang2025visionzip}, SparseVLM~\citep{zhang2024sparsevlm}, DART~\citep{wen2025stop}, DivPrune~\citep{alvar2025divprune} and Zoo-Prune~\citep{kim2026zoo}, as well as training-based methods LLaVA-Mini~\citep{zhang2025llava} and EPIC~\citep{wen2026efficient}. Our main comparisons consider pruning at 20\% and 5\% visual-token retention, with 15\% and 10\% retention results reported in the appendix. For cross-backbone evaluation, we further compare with DART and DivPrune on LLaVA-1.5-7B~\citep{liu2023visual}, Qwen3-VL-30B-A3B~\citep{bai2025qwen3}, and Qwen3.5-4B~\citep{qwen3.5}, and evaluate LLaVA-1.5-13B, LLaVA-v1.6-Mistral-7B~\citep{liu2024llavanext}, and Qwen3-VL-8B in Appendix~\ref{appendix:additional_cross_backbone_results}. All the baselines follow their original pruning schedules when computing the retention rate: the initial full-token layers (layers 0–1) are excluded for FastV, DART, and SparseVLM, while all LLM layers are included for DivPrune, VisionZip, and ZOO-Prune.

\subsection{Main Results}
\paragraph{Results on Single-Image.} 
Table~\ref{tab:baseline_acc_comparison_main} compares \method with both training-free visual token pruning and training-based compression methods on Qwen3-VL-4B. \method achieves an average score of 74.4, retaining 92.9\% of the uncompressed model performance. Compared with the strongest 5\%-retention pruning baseline, DivPrune, \method improves the average score by 9.1 points and achieves higher accuracy across benchmarks. It also outperforms the training-based baselines LLaVA-Mini and EPIC by 9.4 and 7.5 points, respectively. Notably, \method slightly surpasses the best result obtained by pruning methods that retain 20\% of the visual tokens. The recurrent adapter further improves the average score from 74.4 to 75.4 over the embedding adapter. This suggests that the embedding adapter already captures much of the layer-specific visual information, while lightweight cross-layer evolution can further improve the predicted visual memory. These results demonstrate that compressing visual computation along the hidden dimension provides a complementary efficiency axis to token pruning~(we further verify that the two mechanisms can be combined in practice, results with DART and DivPrune are provided in Appendix~\ref{appendix:adapter_combine_with_token_pruning_methods}). Results at 10\% and 15\% retention ratios show the same trend and are reported in Appendix~\ref{appendix:additional_token_retention_results}. 

\paragraph{Efficiency.}
\begin{wraptable}{r}{0.5\textwidth}
    \centering
    \caption{
    Efficiency comparison on Video-MME under 5\% visual-token retention. Total Speedup measures the overall inference speedup, including both prefill and decoding.
    }
    \label{tab:prefill_speed_comparison}

    \scriptsize
    \setlength{\tabcolsep}{2.2pt}
    \renewcommand{\arraystretch}{1.08}

    \begin{tabular}{lcccc}
        \toprule
        \textbf{Method}
        & \textbf{FLOPs}
        & \textbf{Peak Mem.}
        & \textbf{Total}
        & \textbf{Prefill} \\
        & \textbf{(\%)}
        & \textbf{(GB)}
        & \textbf{Speedup}
        & \textbf{Speedup} \\
        \midrule

        \rowcolor{gray!15}
        \multicolumn{5}{c}{\textit{Upper Bound}} \\

        Qwen3-VL-4B
        & 100.00
        & 12.94
        & 1.00
        & 1.00 \\

        \rowcolor{gray!15}
        \multicolumn{5}{c}{\textit{Retain 5\% Visual Tokens}} \\

        FastV
        & 21.49
        & 11.26
        & 1.22
        & 1.41 \\

        DART
        & 21.23
        & 11.12
        & 1.13
        & 1.24 \\

        VisionZip
        & 16.62
        & 10.58
        & 1.30
        & 1.45 \\

        DivPrune
        & 16.58
        & 10.60
        & 1.30
        & 1.47 \\

        ZOO-Prune
        & 23.90
        & 11.76
        & 1.23
        & 1.29 \\

        SparseVLM
        & 21.60
        & 11.27
        & 1.19
        & 1.36 \\

        \midrule

        \rowcolor{myblue}
        \textbf{\method}
        & 17.90
        & 12.97
        & 1.30
        & 1.50 \\

        \bottomrule
    \end{tabular}
\end{wraptable}

Table~\ref{tab:prefill_speed_comparison} compares the efficiency of \method with token pruning methods on Video-MME. \method reduces the computation to 17.90\% of the uncompressed model FLOPs, comparable to the 16.58--23.90\% range of methods retaining only 5\% of visual tokens. In practice, \method achieves a $1.30\times$ end-to-end speedup and a $1.50\times$ prefill speedup over the base model, with prefill efficiency comparable to the strongest pruning baselines while retaining all visual tokens. Importantly, this comparable efficiency is accompanied by substantially better task performance. These results highlight that inference savings can be obtained without shortening the visual sequence. This provides a complementary efficiency mechanism to sequence-length reduction, while retaining access to the complete visual evidence at every layer. We provide additional results for the 20\% visual-token retention setting in Appendix~\ref{appendix:additional efficiency_results}.

\paragraph{Generalization across Backbones.}
\begin{table*}[htbp]
    \centering
    \caption{
    Results of different visual token compression methods across
    various MLLMs. \textit{Vanilla} denotes the uncompressed
    upper bound. DART and DivPrune retain 5\% of the visual tokens.
    }
    \label{tab:different_models_acc_comparison}

    \scriptsize
    \setlength{\tabcolsep}{2.2pt}
    \renewcommand{\arraystretch}{1.05}

    \begin{tabular}{llcccccccccc}
        \toprule
        \textbf{Model}
        & \textbf{Method}
        & \textbf{MMStar}
        & \textbf{RWQA}
        & \textbf{GQA}
        & \textbf{MMB}
        & \textbf{MMB-CN}
        & \textbf{MME}
        & \textbf{POPE}
        & \textbf{SQA}
        & \textbf{VQA-v2}
        & \textbf{Avg.} \\
        \midrule

        \multirow{4}{*}{\textbf{LLaVA-1.5-7B}}
        & Vanilla
        & 37.0 & 54.9 & 61.2 & 72.5 & 67.9
        & 71.7 & 83.4 & 65.6 & 75.9 & 65.6 \\

        & DART
        & 29.3 & 46.5 & 51.6 & 63.8 & 55.7
        & 68.9 & 68.1 & 69.0 & 61.5 & 57.2 \\

        & DivPrune
        & 31.3 & 49.0 & 53.4 & 66.7 & 57.3
        & 66.3 & 79.3 & 64.5 & 70.5 & 59.8 \\

        \rowcolor{myblue}
        & \textbf{\method}
        & 35.4 & 51.4 & 53.8 & 69.9 & 63.7
        & 68.7 & 79.2 & 65.4 & 69.5 & \textbf{61.9} \\

        \midrule

        \multirow{4}{*}{\textbf{Qwen3-VL-30B-A3B}}
        & Vanilla
        & 70.3 & 72.3 & 65.2 & 89.4 & 90.8
        & 89.4 & 87.9 & 96.3 & 82.6 & 82.7 \\

        & DART
        & 46.2 & 52.3 & 50.1 & 77.2 & 77.9
        & 79.2 & 74.3 & 85.9 & 66.7 & 67.8 \\

        & DivPrune
        & 45.6 & 62.9 & 54.9 & 79.3 & 80.8
        & 83.0 & 77.7 & 85.0 & 72.7 & 71.3 \\

        \rowcolor{myblue}
        & \textbf{\method}
        & 62.4 & 65.1 & 62.4 & 86.5 & 86.8
        & 87.0 & 87.8 & 88.6 & 80.7 & \textbf{78.6} \\

        \midrule

        \multirow{4}{*}{\textbf{Qwen3.5-4B}}
        & Vanilla
        & 48.9 & 67.2 & 63.4 & 84.3 & 77.2
        & 84.0 & 88.7 & 78.4 & 80.0 & 74.7 \\

        & DART
        & 30.5 & 48.0 & 46.1 & 67.8 & 61.9
        & 71.2 & 68.3 & 64.3 & 60.8 & 57.7 \\

        & DivPrune
        & 34.8 & 56.0 & 48.8 & 74.2 & 64.0
        & 75.8 & 73.4 & 71.5 & 65.6 & 62.7 \\

        \rowcolor{myblue}
        & \textbf{\method}
        & 40.1 & 65.8 & 60.5 & 81.3 & 75.1
        & 79.7 & 89.0 & 74.8 & 77.7 & \textbf{71.6} \\

        \bottomrule
    \end{tabular}
\end{table*}
We further evaluate \method on 3 additional MLLM backbones from LLaVA and Qwen families, including LLaVA-1.5-7B, Qwen3-VL-30B-A3B, and Qwen3.5-4B. The latter adopts a hybrid attention architecture, for which we provide additional analyses in Appendix~\ref{appendix:linear_attention_analysis}. As shown in Table~\ref{tab:different_models_acc_comparison}, \method consistently preserves a large fraction of the vanilla-model performance across different model scales and architectures. The gains over token-pruning baselines are particularly pronounced on the Qwen family. On Qwen3-VL-30B-A3B, \method reaches an average score of 78.6, compared with 71.3 for DivPrune under 5\% token retention. On Qwen3.5-4B, \method achieves 71.6, improving over DivPrune at 62.7. Similar improvements are also observed on LLaVA-1.5-7B. Across these three backbones, \method retains approximately 94--96\% of the corresponding vanilla-model performance, despite substantial differences in model scale and architecture. These results support the generality of visual-memory prediction as an complementary to explicitly propagating visual tokens through the full Transformer computation. Additional cross-backbone results are provided in Appendix~\ref{appendix:additional_cross_backbone_results},
including comparisons under 20\% visual-token retention across 6 backbones
and 5\% retention results on the three backbones.

\paragraph{Results on Multi-Image and Video.}
\begin{wraptable}{r}{0.55\textwidth}
    \centering
    \caption{
    Results on multi-image and video benchmarks.
    Training-free baselines are evaluated at 5\% visual-token retention.
    }
    \label{tab:video_multi_image_comparison}

    \scriptsize
    \setlength{\tabcolsep}{2.2pt}
    \renewcommand{\arraystretch}{1.08}

    \begin{tabular}{lcccc}
        \toprule
        \textbf{Method}
        & \textbf{MuirBench}
        & \textbf{Video-MME}
        & \textbf{MVBench}
        & \textbf{Avg.} \\
        \midrule

        \rowcolor{gray!15}
        \multicolumn{5}{c}{\textit{Upper Bound}} \\

        Qwen3-VL-4B
        & 53.5 & 52.0 & 61.6 & 55.7 \\

        \rowcolor{gray!10}
        \multicolumn{5}{c}{\textit{Training-Free (Retain 5\% Visual Tokens)}} \\

        FastV
        & 42.1 & 45.3 & 41.2 & 42.9 \\

        DART
        & 45.1 & 51.3 & 53.3 & 49.9 \\

        VisionZip
        & 43.6 & 48.2 & 46.5 & 46.1 \\

        SparseVLM
        & 42.2 & 47.2 & 45.4 & 44.9 \\

        DivPrune
        & 46.1 & 51.3 & 53.3 & 50.2 \\

        ZOO-Prune
        & 43.9 & 47.9 & 48.5 & 46.8 \\

        \rowcolor{gray!10}
        \multicolumn{5}{c}{\textit{Training-Based (Retain 5\% Visual Tokens)}} \\

        LLaVA-Mini
        & 43.8 & 47.5 & 45.2 & 45.5 \\

        EPIC
        & 42.7 & 48.4 & 48.8 & 46.6 \\

        \midrule

        \rowcolor{myblue}
        \textbf{\method} (Sin-Img.)
        & 40.7 & 50.5 & 57.4 & 49.5 \\

        \rowcolor{myblue}
        \textbf{\method} (Mul-Img.)
        & 45.5 & 51.0 & 59.5 & \textbf{52.0} \\

        \bottomrule
    \end{tabular}
    \vspace{-3mm}
\end{wraptable}
As shown in Table~\ref{tab:video_multi_image_comparison}, the embedding adapter trained only on single-image data achieves an average score of 49.5, while incorporating multi-image and video training data~(the training recipe is detailed in Appendix~\ref{appendix:additional_implementation_and_training_details}) improves the average to 52.0. The largest gains appear on MuirBench and MVBench, increasing from 40.7 to 45.5 and from 57.4 to 59.5, respectively, while Video-MME improves from 50.5 to 51.0. The resulting 52.0 average surpasses the strongest 5\%-retention pruning baseline at 50.2. These results indicate the visual-memory formulation remains effective for longer and more complex visual contexts. Additional results with pruning baselines retaining 20\% of the visual tokens are provided in Appendix~\ref{appendix:additional_result_on_multi-Image_and_video}.

\subsection{How Much Visual Computation Is Actually Necessary?}

We further ask how much of the visual computation is actually needed to preserve the model behavior. We study this question at three levels: (1) characterizing the native spectral structure of visual computation, (2) testing whether low-rank visual effects are sufficient to maintain model performance, (3) propagating low-rank interventions sequentially through the network. These analyses progressively examine whether the visual computation used by the model can be represented within a substantially smaller effective subspace.

\paragraph{Spectral Structure of Native Visual Computation.} 

\begin{table}[t]
    \centering
    \caption{
    Average rank statistics of visual-token computations across Transformer layers.
    $r_{95}$ denotes the minimum rank retaining 95\% of the squared singular-value energy,
    and ER denotes effective rank.
    }
    \label{tab:low_rank_statistics}

    \footnotesize

    \setlength{\tabcolsep}{3pt}

    \renewcommand{\arraystretch}{1.25}

    \resizebox{0.6\linewidth}{!}{%
    \begin{tabular}{
        l
        S[table-format=4.1]
        >{\columncolor{gray!7}}S[table-format=3.1]
        S[table-format=3.1]
        S[table-format=3.1]
        >{\columncolor{gray!7}}S[table-format=2.1]
        S[table-format=2.1]
        S[table-format=4.1]
        >{\columncolor{gray!7}}S[table-format=3.1]
        S[table-format=3.1]
    }
        \toprule

        \textbf{Dataset}
        & \multicolumn{3}{c}{$\boldsymbol{Q}$}
        & \multicolumn{3}{c}{$\boldsymbol{QK^\top}$}
        & \multicolumn{3}{c}{\textbf{Attention Output}} \\

        \cmidrule(lr){2-4}
        \cmidrule(lr){5-7}
        \cmidrule(lr){8-10}

        &
        {\textbf{Full}}
        & {$\boldsymbol{r_{95}}$}
        & {\textbf{ER}}
        & {\textbf{Full}}
        & {$\boldsymbol{r_{95}}$}
        & {\textbf{ER}}
        & {\textbf{Full}}
        & {$\boldsymbol{r_{95}}$}
        & {\textbf{ER}} \\

        \midrule

        \rowcolor{qwenpurple}
        \multicolumn{10}{l}{\textbf{Qwen3-VL-4B}} \\
        \addlinespace[1pt]

        MMStar
        & 213.9 & 42.7 & 108.4
        & 119.2 & 5.0 & 19.3
        & 212.2 & 47.0 & 101.2 \\

        RWQA
        & 1296.0 & 112.7 & 530.0
        & 128.0 & 10.7 & 35.3
        & 1296.0 & 103.5 & 378.5 \\

        SQA
        & 170.5 & 35.4 & 89.1
        & 101.3 & 4.1 & 16.0
        & 170.5 & 38.0 & 83.4 \\

        \addlinespace[2pt]
        \midrule
        \addlinespace[1pt]

        \rowcolor{llavaorange}
        \multicolumn{10}{l}{\textbf{LLaVA-1.5-7B}} \\
        \addlinespace[1pt]

        MMStar
        & 576.0 & 79.8 & 243.9
        & 128.0 & 2.3 & 15.6
        & 576.0 & 37.3 & 151.8 \\

        RWQA
        & 576.0 & 81.3 & 251.1
        & 128.0 & 2.3 & 15.6
        & 576.0 & 37.6 & 156.5 \\

        SQA
        & 576.0 & 71.8 & 230.4
        & 128.0 & 2.2 & 14.7
        & 576.0 & 29.6 & 134.5 \\

        \bottomrule
    \end{tabular}%
    }
\end{table}
We first examine the intrinsic dimensionality of visual-token computation in the unmodified model. For each sample and Transformer layer, we analyze three quantities associated with visual tokens: the visual queries after query normalization and RoPE, the per-head visual $QK^\top$ matrices before masking and softmax, and the visual attention outputs after the output projection but before the residual connection. We characterize the spectral concentration using two complementary statistics. $r_{95}$ denotes the minimum rank required to retain 95\% of the squared singular-value energy. We additionally report the effective rank, defined as $\mathrm{ER}=\exp(-\sum_i p_i\log p_i)$, where $p_i=\sigma_i/\sum_j\sigma_j$ is the normalized singular-value spectrum. As shown in Table~\ref{tab:low_rank_statistics}, across both Qwen3-VL-4B and LLaVA-1.5-7B, these quantities exhibit strongly concentrated spectra. In particular, the visual attention outputs require only 38.0--103.5 directions on Qwen3-VL-4B and 29.6--37.6 directions on LLaVA-1.5-7B to retain 95\% of the spectral energy, substantially below their available rank. These results suggest that the full representational capacity allocated to visual processing is not uniformly utilized across directions.

\paragraph{Low-Rank Sufficiency of Visual Information.}
\begin{wrapfigure}{r}{0.68\textwidth}
    \centering
    \includegraphics[
        width=\linewidth,
        trim=5 5 5 5,
        clip
    ]{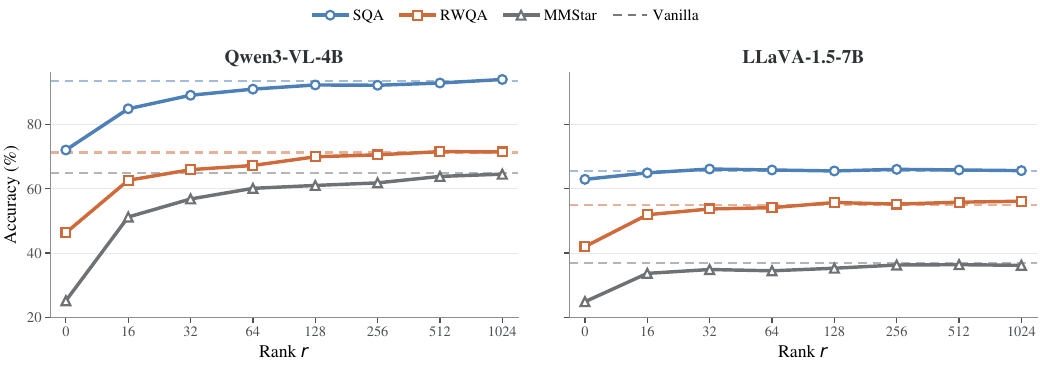}
    \caption{
    Accuracy recovery under low-rank visual attention effect.
    We vary the retained rank $r$ from 0 to 1024.
    }
    \label{fig:low_rank_accuracy_recovery_t2v}
\end{wrapfigure}
To test whether the observed spectral concentration is functionally meaningful, we run the unmodified model to obtain the native hidden-state trajectory and isolate the attention effect induced by visual information flow at each layer. This visual effect is then projected onto a rank-$r$ subspace before being passed to subsequent computation, while the effect itself is always computed from the native trajectory. As shown in Figure~\ref{fig:low_rank_accuracy_recovery_t2v}, performance recovers rapidly as the retained rank increases. On Qwen3-VL-4B, rank 128 reaches 92.2 on SQA, 69.9 on RWQA, and 61.0 on MMStar, compared with 93.4, 71.4, and 64.8 when the full visual effect is retained. A similar trend is observed on LLaVA-1.5-7B, where relatively small ranks already approach the full-effect performance. These results indicate that, given the native model trajectory, most task-relevant visual effects can be represented within a substantially smaller hidden subspace.

\paragraph{Sequential Low-Rank Intervention on Visual Information Flow.}
\begin{table*}[htbp]
    \centering
    \caption{
    Low-rank intervention on visual-to-text information transfer. visual-to-text attention is blocked at selected layers, and only the leading $r$ directions of the induced attention-output difference are restored. For Qwen3-VL-4B, the middle and last 10-layer groups correspond to layers 13--22 and 26--35; for LLaVA-1.5-7B, they correspond to layers 11--20 and 22--31, respectively.
    }
    \label{tab:intervention_layer_low_rank_acc}

    \setlength{\tabcolsep}{4.2pt}
    \renewcommand{\arraystretch}{1.10}

    \begin{adjustbox}{width=\textwidth}
    \begin{tabular}{
        ll
        c
        >{\columncolor{gray!8}}c
        c c
        >{\columncolor{gray!8}}c
        @{\hspace{10pt}}
        c
        >{\columncolor{gray!8}}c
        c c
        >{\columncolor{gray!8}}c
    }
        \toprule

        \multirow{2}{*}{\textbf{Dataset}}
        & \multirow{2}{*}{\textbf{Intervention}}
        & \multicolumn{5}{c}{\textbf{Qwen3-VL-4B}}
        & \multicolumn{5}{c}{\textbf{LLaVA-1.5-7B}} \\

        \cmidrule(lr){3-7}
        \cmidrule(lr){8-12}

        &
        & \textit{Vanilla}
        & $\boldsymbol{r=0}$
        & $\boldsymbol{r=32}$
        & $\boldsymbol{r=64}$
        & $\boldsymbol{r=128}$
        & \textit{Vanilla}
        & $\boldsymbol{r=0}$
        & $\boldsymbol{r=32}$
        & $\boldsymbol{r=64}$
        & $\boldsymbol{r=128}$ \\

        \midrule

        \multirow{5}{*}{\textbf{RWQA}}
        & First 5
        & \multirow{5}{*}{71.2}
        & 72.6 & 71.9 & 71.8 & 71.5
        & \multirow{5}{*}{54.9}
        & 53.2 & 55.8 & 56.2 & 56.0 \\

        & First 10
        &
        & 72.9 & 71.9 & 72.0 & 72.0
        &
        & 49.4 & 54.9 & 55.2 & 55.8 \\

        & Middle 10
        &
        & 45.9 & 69.9 & 69.5 & 70.5
        &
        & 48.1 & 54.3 & 55.6 & 55.2 \\

        & Last 10
        &
        & 71.1 & 71.0 & 71.1 & 71.5
        &
        & 56.0 & 55.0 & 55.2 & 55.2 \\

        & All
        &
        & 46.1 & 68.4 & 69.2 & 70.7
        &
        & 42.4 & 53.2 & 54.9 & 54.8 \\

        \addlinespace[2pt]
        \midrule
        \addlinespace[1pt]

        \multirow{5}{*}{\textbf{MMStar}}
        & First 5
        & \multirow{5}{*}{64.9}
        & 64.8 & 65.0 & 65.4 & 65.2
        & \multirow{5}{*}{37.0}
        & 37.3 & 35.8 & 36.4 & 36.7 \\

        & First 10
        &
        & 62.8 & 64.7 & 64.9 & 65.1
        &
        & 31.0 & 34.8 & 34.9 & 35.6 \\

        & Middle 10
        &
        & 40.5 & 61.0 & 63.3 & 64.4
        &
        & 36.3 & 36.9 & 37.4 & 37.0 \\

        & Last 10
        &
        & 65.2 & 65.1 & 64.8 & 64.9
        &
        & 36.4 & 36.8 & 36.8 & 37.0 \\

        & All
        &
        & 25.6 & 59.5 & 63.0 & 64.6
        &
        & 24.9 & 34.0 & 34.9 & 35.8 \\

        \bottomrule
    \end{tabular}
    \end{adjustbox}
\end{table*}
A stronger question is whether this low-rank sufficiency still holds when the compressed visual effect is propagated through subsequent layers. We further apply the low-rank intervention sequentially across Transformer layers, such that the compressed output at one layer directly affects the hidden states processed by subsequent layers. As shown in Table~\ref{tab:intervention_layer_low_rank_acc}, low-dimensional visual information remains sufficient even under this setting. When all layers of Qwen3-VL-4B are intervened, completely removing direct visual information reduces RWQA from 71.2 to 46.1 and MMStar from 64.9 to 25.6, approaching chance-level performance, whereas restoring only 128 hidden directions recovers the scores to 70.7 and 64.6, respectively. The layer-group interventions further reveal that the dependence on visual information is highly non-uniform across depth. On Qwen3-VL-4B, blocking visual information in the middle layers causes the largest degradation, while the first few and final layers are largely insensitive. In contrast, LLaVA-1.5-7B exhibits a more distributed and task-dependent pattern: on RWQA, both the early and middle layers contribute noticeably, whereas on MMStar the first ten layers are more sensitive than the middle or final layers. These results suggest that the effective dimensionality of visual information remains small even when compression is propagated through the network, while the importance of visual-token processing is highly non-uniform across depth and concentrated in a model-specific subset of Transformer layers. Thus, \method enables flexible layer-wise skipping rather than applying apply full visual-token computation uniformly at every layer in standard MLLMs~(we further validate this in Appendix~\ref{appendix:adapter_layer_skipping})). 

\section{Related Work}
\label{Related Work}

\textbf{Visual Representation Redundancy.}
Visual redundancy exists not only across tokens but also in the hidden representation space. LRCP~\citep{lu2026lrcp} uses PCA to identify dominant visual subspaces and retains tokens that are poorly represented by them. Other works reduce layer-wise visual computation directly. EE-MLLM~\citep{ma2024ee} removes self-attention among visual tokens, while ShortV~\citep{yuan2025shortv} and \citet{luo2026attend} estimate layer importance from changes in the output distribution. Unlike these methods, which remove or preserve parts of the original computation, \method learns an alternative pathway to construct visual states. KV Prediction~\citep{horton2024kv} uses a small auxiliary model to predict the KV cache of a larger model. \method instead predicts visual memory before KV projection with lightweight adapters and reuses the frozen key and value projections.

\textbf{Token Reduction for Efficient MLLMs.}
Visual token reduction lowers MLLM inference cost by selecting, merging, or compressing visual tokens. Training-free methods~\citep{yang2025visionzip, alvar2025divprune, wen2025stop, zhang2024sparsevlm, chen2024image, kim2026zoo} mainly gain efficiency by reducing sequence length, while training-based methods~\citep{zhang2025llava, wen2026efficient} improve adaptation to compact visual representations. Some methods also allow removed tokens to re-enter later layers. SwiftVLM~\citep{qian2026swiftvlm} uses cross-layer bypasses and residual connections to recover unselected tokens. In contrast, \method does not select or remove visual tokens. It constructs key-value context for all visual tokens at every layer. Thus, our method changes how visual states are generated rather than which visual tokens are retained, providing a different efficiency direction from sequence-length compression.

\section{Conclusion}
\label{sec:conclusion}
In this work, we revisit the common assumption that visual tokens undergo full Transformer evolution in MLLMs. Our analysis shows that layer-wise visual representations exhibit substantial redundancy in the hidden-channel dimension and predictability across depth, suggesting that useful visual context can be preserved without repeatedly applying the full visual Transformer pathway. Motivated by these observations, we introduce \method, which replaces visual-token evolution with lightweight low-rank adapters while preserving all visual tokens for text retrieval. Across multiple MLLM backbones and image, multi-image, and video benchmarks, \method achieves a favorable accuracy--efficiency trade-off relative to visual token pruning, while operating in a comparable computational regime. These results suggest that reducing multimodal inference cost does not necessarily require removing visual tokens. Efficiently constructing visual states provides a complementary efficiency axis to sequence-length reduction, opening a another direction for scalable MLLMs.

\bibliographystyle{plainnat}
\bibliography{references}

@article{luo2026attend,
  author       = {Zhaoyang Luo and
                  Runmin Dong and
                  Miao Yang and
                  Fan Wei and
                  Yushan Lai and
                  Bin Luo and
                  Haohuan Fu},
  title        = {Attend, Transform, or Silence: Operator-Level Visual Skipping for
                  Efficient Multimodal {LLM} Inference},
  journal      = {CoRR},
  volume       = {abs/2606.31903},
  year         = {2026},
  url          = {https://doi.org/10.48550/arXiv.2606.31903},
  doi          = {10.48550/ARXIV.2606.31903},
  eprinttype   = {arXiv},
  eprint       = {2606.31903},
  bibsource    = {dblp computer science bibliography, https://dblp.org}
}

@article{qian2026swiftvlm,
  author       = {Chen Qian and
                  Xinran Yu and
                  Danyang Li and
                  Guoxuan Chi and
                  Zheng Yang and
                  Qiang Ma and
                  Xin Miao},
  title        = {SwiftVLM: Efficient Vision-Language Model Inference via Cross-Layer
                  Token Bypass},
  journal      = {CoRR},
  volume       = {abs/2602.03134},
  year         = {2026},
  url          = {https://doi.org/10.48550/arXiv.2602.03134},
  doi          = {10.48550/ARXIV.2602.03134},
  eprinttype   = {arXiv},
  eprint       = {2602.03134},
  bibsource    = {dblp computer science bibliography, https://dblp.org}
}

@article{lu2026lrcp,
  author       = {Hongyu Lu and
                  Feng Zhang and
                  Wenwei Jin and
                  Huanling Hu and
                  Tianjun Shi and
                  Shikai Jiang and
                  Yao Hu and
                  Jiawei Li},
  title        = {{LRCP:} Low-Rank Compressibility Guided Visual Token Pruning for Efficient
                  LVLMs},
  journal      = {CoRR},
  volume       = {abs/2605.15621},
  year         = {2026},
  url          = {https://doi.org/10.48550/arXiv.2605.15621},
  doi          = {10.48550/ARXIV.2605.15621},
  eprinttype   = {arXiv},
  eprint       = {2605.15621},
  bibsource    = {dblp computer science bibliography, https://dblp.org}
}

@article{ma2024ee,
  author       = {Feipeng Ma and
                  Yizhou Zhou and
                  Hebei Li and
                  Zilong He and
                  Siying Wu and
                  Fengyun Rao and
                  Yueyi Zhang and
                  Xiaoyan Sun},
  title        = {{EE-MLLM:} {A} Data-Efficient and Compute-Efficient Multimodal Large
                  Language Model},
  journal      = {CoRR},
  volume       = {abs/2408.11795},
  year         = {2024},
  url          = {https://doi.org/10.48550/arXiv.2408.11795},
  doi          = {10.48550/ARXIV.2408.11795},
  eprinttype   = {arXiv},
  eprint       = {2408.11795},
  bibsource    = {dblp computer science bibliography, https://dblp.org}
}

@inproceedings{yuan2025shortv,
  author       = {Qianhao Yuan and
                  Qingyu Zhang and
                  Yanjiang Liu and
                  Jiawei Chen and
                  Yaojie Lu and
                  Hongyu Lin and
                  Jia Zheng and
                  Xianpei Han and
                  Le Sun},
  title        = {ShortV: Efficient Multimodal Large Language Models by Freezing Visual
                  Tokens in Ineffective Layers},
  booktitle    = {{IEEE/CVF} International Conference on Computer Vision, {ICCV} 2025,
                  Honolulu, HI, USA, October 19-25, 2025},
  pages        = {329--339},
  publisher    = {{IEEE}},
  year         = {2025},
  url          = {https://doi.org/10.1109/ICCV51701.2025.00038},
  doi          = {10.1109/ICCV51701.2025.00038},
  bibsource    = {dblp computer science bibliography, https://dblp.org}
}

@inproceedings{alvar2025divprune,
  author       = {Saeed Ranjbar Alvar and
                  Gursimran Singh and
                  Mohammad Akbari and
                  Yong Zhang},
  title        = {DivPrune: Diversity-based Visual Token Pruning for Large Multimodal
                  Models},
  booktitle    = {{IEEE/CVF} Conference on Computer Vision and Pattern Recognition,
                  {CVPR} 2025, Nashville, TN, USA, June 11-15, 2025},
  pages        = {9392--9401},
  publisher    = {Computer Vision Foundation / {IEEE}},
  year         = {2025},
  url          = {https://openaccess.thecvf.com/content/CVPR2025/html/Alvar\_DivPrune\_Diversity-based\_Visual\_Token\_Pruning\_for\_Large\_Multimodal\_Models\_CVPR\_2025\_paper.html},
  doi          = {10.1109/CVPR52734.2025.00877},
  bibsource    = {dblp computer science bibliography, https://dblp.org}
}

@inproceedings{yang2025visionzip,
  author       = {Senqiao Yang and
                  Yukang Chen and
                  Zhuotao Tian and
                  Chengyao Wang and
                  Jingyao Li and
                  Bei Yu and
                  Jiaya Jia},
  title        = {VisionZip: Longer is Better but Not Necessary in Vision Language Models},
  booktitle    = {{IEEE/CVF} Conference on Computer Vision and Pattern Recognition,
                  {CVPR} 2025, Nashville, TN, USA, June 11-15, 2025},
  pages        = {19792--19802},
  publisher    = {Computer Vision Foundation / {IEEE}},
  year         = {2025},
  url          = {https://openaccess.thecvf.com/content/CVPR2025/html/Yang\_VisionZip\_Longer\_is\_Better\_but\_Not\_Necessary\_in\_Vision\_Language\_CVPR\_2025\_paper.html},
  doi          = {10.1109/CVPR52734.2025.01843},
  bibsource    = {dblp computer science bibliography, https://dblp.org}
}

@inproceedings{zhang2024sparsevlm,
  author       = {Yuan Zhang and
                  Chun{-}Kai Fan and
                  Junpeng Ma and
                  Wenzhao Zheng and
                  Tao Huang and
                  Kuan Cheng and
                  Denis A. Gudovskiy and
                  Tomoyuki Okuno and
                  Yohei Nakata and
                  Kurt Keutzer and
                  Shanghang Zhang},
  title        = {SparseVLM: Visual Token Sparsification for Efficient Vision-Language
                  Model Inference},
  booktitle    = {Forty-second International Conference on Machine Learning, {ICML}
                  2025, Vancouver, BC, Canada, July 13-19, 2025},
  series       = {Proceedings of Machine Learning Research},
  volume       = {267},
  publisher    = {{PMLR} / OpenReview.net},
  year         = {2025},
  url          = {https://proceedings.mlr.press/v267/zhang25s.html},
  bibsource    = {dblp computer science bibliography, https://dblp.org}
}

@inproceedings{wen2025stop,
  author       = {Zichen Wen and
                  Yifeng Gao and
                  Shaobo Wang and
                  Junyuan Zhang and
                  Qintong Zhang and
                  Weijia Li and
                  Conghui He and
                  Linfeng Zhang},
  editor       = {Christos Christodoulopoulos and
                  Tanmoy Chakraborty and
                  Carolyn Rose and
                  Violet Peng},
  title        = {Stop Looking for "Important Tokens" in Multimodal Language
                  Models: Duplication Matters More},
  booktitle    = {Proceedings of the 2025 Conference on Empirical Methods in Natural
                  Language Processing, {EMNLP} 2025, Suzhou, China, November 4-9, 2025},
  pages        = {9961--9980},
  publisher    = {Association for Computational Linguistics},
  year         = {2025},
  url          = {https://doi.org/10.18653/v1/2025.emnlp-main.505},
  doi          = {10.18653/V1/2025.EMNLP-MAIN.505},
  bibsource    = {dblp computer science bibliography, https://dblp.org}
}

@inproceedings{kim2026zoo,
  author       = {Youngeun Kim and
                  Youjia Zhang and
                  Huiling Liu and
                  Aecheon Jung and
                  Sunwoo Lee and
                  Sungeun Hong},
  title        = {ZOO-Prune: Training-Free Token Pruning via Zeroth-Order Gradient Estimation
                  in Vision-Language Models},
  booktitle    = {{IEEE/CVF} Conference on Computer Vision and Pattern Recognition,
                  {CVPR} 2026, Denver, CO, USA, June 3-7, 2026},
  pages        = {39572--39582},
  publisher    = {Computer Vision Foundation},
  year         = {2026},
  url          = {https://openaccess.thecvf.com/content/CVPR2026/html/Kim\_ZOO-Prune\_Training-Free\_Token\_Pruning\_via\_Zeroth-Order\_Gradient\_Estimation\_in\_Vision-Language\_CVPR\_2026\_paper.html},
  bibsource    = {dblp computer science bibliography, https://dblp.org}
}

@inproceedings{chen2024image,
  author       = {Liang Chen and
                  Haozhe Zhao and
                  Tianyu Liu and
                  Shuai Bai and
                  Junyang Lin and
                  Chang Zhou and
                  Baobao Chang},
  editor       = {Ales Leonardis and
                  Elisa Ricci and
                  Stefan Roth and
                  Olga Russakovsky and
                  Torsten Sattler and
                  G{\"{u}}l Varol},
  title        = {An Image is Worth 1/2 Tokens After Layer 2: Plug-and-Play Inference
                  Acceleration for Large Vision-Language Models},
  booktitle    = {Computer Vision - {ECCV} 2024 - 18th European Conference, Milan, Italy,
                  September 29-October 4, 2024, Proceedings, Part {LXXXI}},
  series       = {Lecture Notes in Computer Science},
  volume       = {15139},
  pages        = {19--35},
  publisher    = {Springer},
  year         = {2024},
  url          = {https://doi.org/10.1007/978-3-031-73004-7\_2},
  doi          = {10.1007/978-3-031-73004-7\_2},
  bibsource    = {dblp computer science bibliography, https://dblp.org}
}

@inproceedings{wen2026efficient,
  author       = {Zichen Wen and
                  Shaobo Wang and
                  Yufa Zhou and
                  Junyuan Zhang and
                  Qintong Zhang and
                  Yifeng Gao and
                  Zhaorun Chen and
                  Bin Wang and
                  Weijia Li and
                  Conghui He and
                  Linfeng Zhang},
  title        = {Efficient Multi-modal Large Language Models via Progressive Consistency
                  Distillation},
  booktitle    = {Advances in Neural Information Processing Systems 38: Annual Conference
                  on Neural Information Processing Systems 2025, NeurIPS 2025, San Diego,
                  CA, USA, December 2-7, 2025 / Mexico City, Mexico, November 30 - December
                  5, 2025},
  year         = {2025},
  url          = {http://papers.nips.cc/paper\_files/paper/2025/hash/6518f9339196e172fa0ceef48a85543a-Abstract-Conference.html},
  bibsource    = {dblp computer science bibliography, https://dblp.org}
}

@inproceedings{zhang2025llava,
  author       = {Shaolei Zhang and
                  Qingkai Fang and
                  Zhe Yang and
                  Yang Feng},
  title        = {LLaVA-Mini: Efficient Image and Video Large Multimodal Models with
                  One Vision Token},
  booktitle    = {The Thirteenth International Conference on Learning Representations,
                  {ICLR} 2025, Singapore, April 24-28, 2025},
  publisher    = {OpenReview.net},
  year         = {2025},
  url          = {https://openreview.net/forum?id=UQJ7CDW8nb},
  bibsource    = {dblp computer science bibliography, https://dblp.org}
}

@inproceedings{agarwal2024policy,
  author       = {Rishabh Agarwal and
                  Nino Vieillard and
                  Yongchao Zhou and
                  Piotr Stanczyk and
                  Sabela Ramos Garea and
                  Matthieu Geist and
                  Olivier Bachem},
  title        = {On-Policy Distillation of Language Models: Learning from Self-Generated
                  Mistakes},
  booktitle    = {The Twelfth International Conference on Learning Representations,
                  {ICLR} 2024, Vienna, Austria, May 7-11, 2024},
  publisher    = {OpenReview.net},
  year         = {2024},
  url          = {https://openreview.net/forum?id=3zKtaqxLhW},
  bibsource    = {dblp computer science bibliography, https://dblp.org}
}

@article{hurst2024gpt,
  author       = {OpenAI},
  title        = {GPT-4o System Card},
  journal      = {CoRR},
  volume       = {abs/2410.21276},
  year         = {2024},
  url          = {https://doi.org/10.48550/arXiv.2410.21276},
  doi          = {10.48550/ARXIV.2410.21276},
  eprinttype   = {arXiv},
  eprint       = {2410.21276},
  bibsource    = {dblp computer science bibliography, https://dblp.org}
}

@article{bai2025qwen3,
  author       = {{Qwen Team}},
  title        = {Qwen3-VL Technical Report},
  journal      = {CoRR},
  volume       = {abs/2511.21631},
  year         = {2025},
  url          = {https://doi.org/10.48550/arXiv.2511.21631},
  doi          = {10.48550/ARXIV.2511.21631},
  eprinttype   = {arXiv},
  eprint       = {2511.21631},
  bibsource    = {dblp computer science bibliography, https://dblp.org}
}

@article{team2025kimi,
  author       = {{Kimi Team}},
  title        = {Kimi-VL Technical Report},
  journal      = {CoRR},
  volume       = {abs/2504.07491},
  year         = {2025},
  url          = {https://doi.org/10.48550/arXiv.2504.07491},
  doi          = {10.48550/ARXIV.2504.07491},
  eprinttype   = {arXiv},
  eprint       = {2504.07491},
  bibsource    = {dblp computer science bibliography, https://dblp.org}
}

@article{hong2026glm,
  author       = {GLM},
  title        = {GLM-5V-Turbo: Toward a Native Foundation Model for Multimodal Agents},
  journal      = {CoRR},
  volume       = {abs/2604.26752},
  year         = {2026},
  url          = {https://doi.org/10.48550/arXiv.2604.26752},
  doi          = {10.48550/ARXIV.2604.26752},
  eprinttype   = {arXiv},
  eprint       = {2604.26752},
  bibsource    = {dblp computer science bibliography, https://dblp.org}
}

@inproceedings{liu2023visual,
  author       = {Haotian Liu and
                  Chunyuan Li and
                  Yuheng Li and
                  Yong Jae Lee},
  title        = {Improved Baselines with Visual Instruction Tuning},
  booktitle    = {{IEEE/CVF} Conference on Computer Vision and Pattern Recognition,
                  {CVPR} 2024, Seattle, WA, USA, June 16-22, 2024},
  pages        = {26286--26296},
  publisher    = {{IEEE}},
  year         = {2024},
  url          = {https://doi.org/10.1109/CVPR52733.2024.02484},
  doi          = {10.1109/CVPR52733.2024.02484},
  bibsource    = {dblp computer science bibliography, https://dblp.org}
}

@inproceedings{chen2024we,
  author       = {Lin Chen and
                  Jinsong Li and
                  Xiaoyi Dong and
                  Pan Zhang and
                  Yuhang Zang and
                  Zehui Chen and
                  Haodong Duan and
                  Jiaqi Wang and
                  Yu Qiao and
                  Dahua Lin and
                  Feng Zhao},
  editor       = {Amir Globersons and
                  Lester Mackey and
                  Danielle Belgrave and
                  Angela Fan and
                  Ulrich Paquet and
                  Jakub M. Tomczak and
                  Cheng Zhang},
  title        = {Are We on the Right Way for Evaluating Large Vision-Language Models?},
  booktitle    = {Advances in Neural Information Processing Systems 37: Annual Conference
                  on Neural Information Processing Systems 2024, NeurIPS 2024, Vancouver,
                  BC, Canada, December 10 - 15, 2024},
  year         = {2024},
  url          = {http://papers.nips.cc/paper\_files/paper/2024/hash/2f8ee6a3d766b426d2618e555b5aeb39-Abstract-Conference.html},
  bibsource    = {dblp computer science bibliography, https://dblp.org}
}

@inproceedings{liu2024mmbench,
  author       = {Yuan Liu and
                  Haodong Duan and
                  Yuanhan Zhang and
                  Bo Li and
                  Songyang Zhang and
                  Wangbo Zhao and
                  Yike Yuan and
                  Jiaqi Wang and
                  Conghui He and
                  Ziwei Liu and
                  Kai Chen and
                  Dahua Lin},
  editor       = {Ales Leonardis and
                  Elisa Ricci and
                  Stefan Roth and
                  Olga Russakovsky and
                  Torsten Sattler and
                  G{\"{u}}l Varol},
  title        = {MMBench: Is Your Multi-modal Model an All-Around Player?},
  booktitle    = {Computer Vision - {ECCV} 2024 - 18th European Conference, Milan, Italy,
                  September 29-October 4, 2024, Proceedings, Part {VI}},
  series       = {Lecture Notes in Computer Science},
  volume       = {15064},
  pages        = {216--233},
  publisher    = {Springer},
  year         = {2024},
  url          = {https://doi.org/10.1007/978-3-031-72658-3\_13},
  doi          = {10.1007/978-3-031-72658-3\_13},
  bibsource    = {dblp computer science bibliography, https://dblp.org}
}

@inproceedings{fu2026mme,
  author       = {Chaoyou Fu and
                  Peixian Chen and
                  Yunhang Shen and
                  Yulei Qin and
                  Mengdan Zhang and
                  Xu Lin and
                  Jinrui Yang and
                  Xiawu Zheng and
                  Ke Li and
                  Xing Sun and
                  Yunsheng Wu and
                  Rongrong Ji and
                  Caifeng Shan and
                  Ran He},
  editor       = {Danielle Belgrave and
                  Cheng Zhang and
                  Laura N. Montoya and
                  Hsuan{-}Tien Lin and
                  Razvan Pascanu and
                  Piotr Koniusz and
                  Marzyeh Ghassemi and
                  Nancy Chen and
                  Iv{\'{a}}n Vladimir Meza Ru{\'{\i}}z and
                  Arturo Loaiza{-}Bonilla},
  title        = {{MME:} {A} Comprehensive Evaluation Benchmark for Multimodal Large
                  Language Models},
  booktitle    = {Advances in Neural Information Processing Systems 38: Annual Conference
                  on Neural Information Processing Systems 2025, NeurIPS 2025, San Diego,
                  CA, USA, December 2-7, 2025 / Mexico City, Mexico, November 30 - December
                  5, 2025},
  year         = {2025},
  url          = {http://papers.nips.cc/paper\_files/paper/2025/hash/d79a27cf2772fe00be7f341efc0eb517-Abstract-Datasets\_and\_Benchmarks\_Track.html},
  bibsource    = {dblp computer science bibliography, https://dblp.org}
}

@inproceedings{pope,
  author       = {Yifan Li and
                  Yifan Du and
                  Kun Zhou and
                  Jinpeng Wang and
                  Wayne Xin Zhao and
                  Ji{-}Rong Wen},
  editor       = {Houda Bouamor and
                  Juan Pino and
                  Kalika Bali},
  title        = {Evaluating Object Hallucination in Large Vision-Language Models},
  booktitle    = {Proceedings of the 2023 Conference on Empirical Methods in Natural
                  Language Processing, {EMNLP} 2023, Singapore, December 6-10, 2023},
  pages        = {292--305},
  publisher    = {Association for Computational Linguistics},
  year         = {2023},
  url          = {https://doi.org/10.18653/v1/2023.emnlp-main.20},
  doi          = {10.18653/V1/2023.EMNLP-MAIN.20},
  bibsource    = {dblp computer science bibliography, https://dblp.org}
}

@inproceedings{lu2022learn,
  author       = {Pan Lu and
                  Swaroop Mishra and
                  Tanglin Xia and
                  Liang Qiu and
                  Kai{-}Wei Chang and
                  Song{-}Chun Zhu and
                  Oyvind Tafjord and
                  Peter Clark and
                  Ashwin Kalyan},
  editor       = {Sanmi Koyejo and
                  S. Mohamed and
                  A. Agarwal and
                  Danielle Belgrave and
                  K. Cho and
                  A. Oh},
  title        = {Learn to Explain: Multimodal Reasoning via Thought Chains for Science
                  Question Answering},
  booktitle    = {Advances in Neural Information Processing Systems 35: Annual Conference
                  on Neural Information Processing Systems 2022, NeurIPS 2022, New Orleans,
                  LA, USA, November 28 - December 9, 2022},
  year         = {2022},
  url          = {http://papers.nips.cc/paper\_files/paper/2022/hash/11332b6b6cf4485b84afadb1352d3a9a-Abstract-Conference.html},
  bibsource    = {dblp computer science bibliography, https://dblp.org}
}

@article{goyal2017making,
  author       = {Yash Goyal and
                  Tejas Khot and
                  Aishwarya Agrawal and
                  Douglas Summers{-}Stay and
                  Dhruv Batra and
                  Devi Parikh},
  title        = {Making the {V} in {VQA} Matter: Elevating the Role of Image Understanding
                  in Visual Question Answering},
  journal      = {Int. J. Comput. Vis.},
  volume       = {127},
  number       = {4},
  pages        = {398--414},
  year         = {2019},
  url          = {https://doi.org/10.1007/s11263-018-1116-0},
  doi          = {10.1007/S11263-018-1116-0},
  bibsource    = {dblp computer science bibliography, https://dblp.org}
}

@inproceedings{fu2025video,
  author       = {Chaoyou Fu and
                  Yuhan Dai and
                  Yongdong Luo and
                  Lei Li and
                  Shuhuai Ren and
                  Renrui Zhang and
                  Zihan Wang and
                  Chenyu Zhou and
                  Yunhang Shen and
                  Mengdan Zhang and
                  Peixian Chen and
                  Yanwei Li and
                  Shaohui Lin and
                  Sirui Zhao and
                  Ke Li and
                  Tong Xu and
                  Xiawu Zheng and
                  Enhong Chen and
                  Caifeng Shan and
                  Ran He and
                  Xing Sun},
  title        = {Video-MME: The First-Ever Comprehensive Evaluation Benchmark of Multi-modal
                  LLMs in Video Analysis},
  booktitle    = {{IEEE/CVF} Conference on Computer Vision and Pattern Recognition,
                  {CVPR} 2025, Nashville, TN, USA, June 11-15, 2025},
  pages        = {24108--24118},
  publisher    = {Computer Vision Foundation / {IEEE}},
  year         = {2025},
  url          = {https://openaccess.thecvf.com/content/CVPR2025/html/Fu\_Video-MME\_The\_First-Ever\_Comprehensive\_Evaluation\_Benchmark\_of\_Multi-modal\_LLMs\_in\_CVPR\_2025\_paper.html},
  doi          = {10.1109/CVPR52734.2025.02245},
  bibsource    = {dblp computer science bibliography, https://dblp.org}
}

@inproceedings{li2024mvbench,
  author       = {Kunchang Li and
                  Yali Wang and
                  Yinan He and
                  Yizhuo Li and
                  Yi Wang and
                  Yi Liu and
                  Zun Wang and
                  Jilan Xu and
                  Guo Chen and
                  Ping Lou and
                  Limin Wang and
                  Yu Qiao},
  title        = {MVBench: {A} Comprehensive Multi-modal Video Understanding Benchmark},
  booktitle    = {{IEEE/CVF} Conference on Computer Vision and Pattern Recognition,
                  {CVPR} 2024, Seattle, WA, USA, June 16-22, 2024},
  pages        = {22195--22206},
  publisher    = {{IEEE}},
  year         = {2024},
  url          = {https://doi.org/10.1109/CVPR52733.2024.02095},
  doi          = {10.1109/CVPR52733.2024.02095},
  bibsource    = {dblp computer science bibliography, https://dblp.org}
}

@inproceedings{wang2025muirbench,
  author       = {Fei Wang and
                  Xingyu Fu and
                  James Y. Huang and
                  Zekun Li and
                  Qin Liu and
                  Xiaogeng Liu and
                  Mingyu Derek Ma and
                  Nan Xu and
                  Wenxuan Zhou and
                  Kai Zhang and
                  Tianyi Lorena Yan and
                  Wenjie Jacky Mo and
                  Hsiang{-}Hui Liu and
                  Pan Lu and
                  Chunyuan Li and
                  Chaowei Xiao and
                  Kai{-}Wei Chang and
                  Dan Roth and
                  Sheng Zhang and
                  Hoifung Poon and
                  Muhao Chen},
  title        = {MuirBench: {A} Comprehensive Benchmark for Robust Multi-image Understanding},
  booktitle    = {The Thirteenth International Conference on Learning Representations,
                  {ICLR} 2025, Singapore, April 24-28, 2025},
  publisher    = {OpenReview.net},
  year         = {2025},
  url          = {https://openreview.net/forum?id=TrVYEZtSQH},
  bibsource    = {dblp computer science bibliography, https://dblp.org}
}

@misc{realworldqa2024,
  title        = {RealWorldQA: A Benchmark for Real-World Spatial Understanding},
  author       = {{xAI}},
  year         = {2024},
  howpublished = {\url{https://huggingface.co/datasets/xai-org/RealworldQA}},
  note         = {Accessed: 2026-09-21}
}

@misc{qwen3.5,
    title  = {{Qwen3.5}: Towards Native Multimodal Agents},
    author = {{Qwen Team}},
    month  = {February},
    year   = {2026},
    url    = {https://qwen.ai/blog?id=qwen3.5}
}

@article{horton2024kv,
  author       = {Maxwell Horton and
                  Qingqing Cao and
                  Chenfan Sun and
                  Yanzi Jin and
                  Sachin Mehta and
                  Mohammad Rastegari and
                  Moin Nabi},
  title        = {{KV} Prediction for Improved Time to First Token},
  journal      = {CoRR},
  volume       = {abs/2410.08391},
  year         = {2024},
  url          = {https://doi.org/10.48550/arXiv.2410.08391},
  doi          = {10.48550/ARXIV.2410.08391},
  eprinttype   = {arXiv},
  eprint       = {2410.08391},
  bibsource    = {dblp computer science bibliography, https://dblp.org}
}

@inproceedings{alayrac2022flamingo,
  author       = {Jean{-}Baptiste Alayrac and
                  Jeff Donahue and
                  Pauline Luc and
                  Antoine Miech and
                  Iain Barr and
                  Yana Hasson and
                  Karel Lenc and
                  Arthur Mensch and
                  Katherine Millican and
                  Malcolm Reynolds and
                  Roman Ring and
                  Eliza Rutherford and
                  Serkan Cabi and
                  Tengda Han and
                  Zhitao Gong and
                  Sina Samangooei and
                  Marianne Monteiro and
                  Jacob L. Menick and
                  Sebastian Borgeaud and
                  Andy Brock and
                  Aida Nematzadeh and
                  Sahand Sharifzadeh and
                  Mikolaj Binkowski and
                  Ricardo Barreira and
                  Oriol Vinyals and
                  Andrew Zisserman and
                  Kar{\'{e}}n Simonyan},
  editor       = {Sanmi Koyejo and
                  S. Mohamed and
                  A. Agarwal and
                  Danielle Belgrave and
                  K. Cho and
                  A. Oh},
  title        = {Flamingo: a Visual Language Model for Few-Shot Learning},
  booktitle    = {Advances in Neural Information Processing Systems 35: Annual Conference
                  on Neural Information Processing Systems 2022, NeurIPS 2022, New Orleans,
                  LA, USA, November 28 - December 9, 2022},
  year         = {2022},
  url          = {http://papers.nips.cc/paper\_files/paper/2022/hash/960a172bc7fbf0177ccccbb411a7d800-Abstract-Conference.html},
  bibsource    = {dblp computer science bibliography, https://dblp.org}
}

@inproceedings{yue2024mmmu,
  author       = {Xiang Yue and
                  Yuansheng Ni and
                  Tianyu Zheng and
                  Kai Zhang and
                  Ruoqi Liu and
                  Ge Zhang and
                  Samuel Stevens and
                  Dongfu Jiang and
                  Weiming Ren and
                  Yuxuan Sun and
                  Cong Wei and
                  Botao Yu and
                  Ruibin Yuan and
                  Renliang Sun and
                  Ming Yin and
                  Boyuan Zheng and
                  Zhenzhu Yang and
                  Yibo Liu and
                  Wenhao Huang and
                  Huan Sun and
                  Yu Su and
                  Wenhu Chen},
  title        = {{MMMU:} {A} Massive Multi-Discipline Multimodal Understanding and
                  Reasoning Benchmark for Expert {AGI}},
  booktitle    = {{IEEE/CVF} Conference on Computer Vision and Pattern Recognition,
                  {CVPR} 2024, Seattle, WA, USA, June 16-22, 2024},
  pages        = {9556--9567},
  publisher    = {{IEEE}},
  year         = {2024},
  url          = {https://doi.org/10.1109/CVPR52733.2024.00913},
  doi          = {10.1109/CVPR52733.2024.00913},
  bibsource    = {dblp computer science bibliography, https://dblp.org}
}

@article{chen2026one,
  author       = {Yongru Chen and
                  Kai Zhang and
                  Zeliang Zong and
                  Yuchen Lu and
                  Wenming Tan and
                  Ye Ren and
                  Jilin Hu},
  title        = {One Layer's Trash is Another Layer's Treasure: Adaptive
                  Layer-wise Visual Token Selection in LVLMs},
  journal      = {CoRR},
  volume       = {abs/2606.14277},
  year         = {2026},
  url          = {https://doi.org/10.48550/arXiv.2606.14277},
  doi          = {10.48550/ARXIV.2606.14277},
  eprinttype   = {arXiv},
  eprint       = {2606.14277},
  bibsource    = {dblp computer science bibliography, https://dblp.org}
}

@article{yang2026reroute,
  author       = {Cheng{-}Yu Yang and
                  Shao{-}Yuan Lo and
                  Yu{-}Lun Liu},
  title        = {Reroute, Don't Remove: Recoverable Visual Token Routing for Vision-Language
                  Models},
  journal      = {CoRR},
  volume       = {abs/2606.12412},
  year         = {2026},
  url          = {https://doi.org/10.48550/arXiv.2606.12412},
  doi          = {10.48550/ARXIV.2606.12412},
  eprinttype   = {arXiv},
  eprint       = {2606.12412},
  bibsource    = {dblp computer science bibliography, https://dblp.org}
}

@inproceedings{shang2025llava,
  author       = {Yuzhang Shang and
                  Mu Cai and
                  Bingxin Xu and
                  Yong Jae Lee and
                  Yan Yan},
  title        = {LLaVA-Prumerge: Adaptive Token Reduction for Efficient Large Multimodal
                  Models},
  booktitle    = {{IEEE/CVF} International Conference on Computer Vision, {ICCV} 2025,
                  Honolulu, HI, USA, October 19-25, 2025},
  pages        = {22857--22867},
  publisher    = {{IEEE}},
  year         = {2025},
  url          = {https://doi.org/10.1109/ICCV51701.2025.02122},
  doi          = {10.1109/ICCV51701.2025.02122},
  bibsource    = {dblp computer science bibliography, https://dblp.org}
}

@inproceedings{gqa,
  author       = {Drew A. Hudson and
                  Christopher D. Manning},
  title        = {{GQA:} {A} New Dataset for Real-World Visual Reasoning and Compositional
                  Question Answering},
  booktitle    = {{IEEE} Conference on Computer Vision and Pattern Recognition, {CVPR}
                  2019, Long Beach, CA, USA, June 16-20, 2019},
  pages        = {6700--6709},
  publisher    = {Computer Vision Foundation / {IEEE}},
  year         = {2019},
  url          = {http://openaccess.thecvf.com/content\_CVPR\_2019/html/Hudson\_GQA\_A\_New\_Dataset\_for\_Real-World\_Visual\_Reasoning\_and\_Compositional\_CVPR\_2019\_paper.html},
  doi          = {10.1109/CVPR.2019.00686},
  bibsource    = {dblp computer science bibliography, https://dblp.org}
}

@misc{liu2024llavanext,
    title={LLaVA-NeXT: Improved reasoning, OCR, and world knowledge},
    url={https://llava-vl.github.io/blog/2024-01-30-llava-next/},
    author={Liu, Haotian and Li, Chunyuan and Li, Yuheng and Li, Bo and Zhang, Yuanhan and Shen, Sheng and Lee, Yong Jae},
    month={January},
    year={2024}
}

@misc{allenai_pixmo_ask_model_anything,
  author       = {{Allen Institute for AI}},
  title        = {PixMo-AskModelAnything},
  howpublished = {Hugging Face Dataset},
  url          = {https://huggingface.co/datasets/allenai/pixmo-ask-model-anything},
  note         = {Accessed: 2026-09-25}
}

@misc{allenai_molmo2_multiimageqa,
  author       = {{Allen Institute for AI}},
  title        = {Molmo2-MultiImageQA},
  howpublished = {Hugging Face Dataset},
  url          = {https://huggingface.co/datasets/allenai/Molmo2-MultiImageQA},
  note         = {Accessed: 2026-09-25}
}

@inproceedings{li2025llava_interleave,
  author       = {Feng Li and
                  Renrui Zhang and
                  Hao Zhang and
                  Yuanhan Zhang and
                  Bo Li and
                  Wei Li and
                  Zejun Ma and
                  Chunyuan Li},
  title        = {LLaVA-Interleave: Tackling Multi-image, Video, and 3D in Large Multimodal
                  Models},
  booktitle    = {The Thirteenth International Conference on Learning Representations,
                  {ICLR} 2025, Singapore, April 24-28, 2025},
  year         = {2025},
  url          = {https://openreview.net/forum?id=oSQiao9GqB},
  bibsource    = {dblp computer science bibliography, https://dblp.org}
}

@inproceedings{feng2025videor1,
  author       = {Kaituo Feng and
                  Kaixiong Gong and
                  Bohao Li and
                  Zonghao Guo and
                  Yibing Wang and
                  Tianshuo Peng and
                  Junfei Wu and
                  Xiaoying Zhang and
                  Benyou Wang and
                  Xiangyu Yue},
  editor       = {Danielle Belgrave and
                  Cheng Zhang and
                  Laura N. Montoya and
                  Hsuan{-}Tien Lin and
                  Razvan Pascanu and
                  Piotr Koniusz and
                  Marzyeh Ghassemi and
                  Nancy Chen and
                  Iv{\'{a}}n Vladimir Meza Ru{\'{\i}}z and
                  Arturo Loaiza{-}Bonilla},
  title        = {Video-R1: Reinforcing Video Reasoning in MLLMs},
  booktitle    = {Advances in Neural Information Processing Systems 38: Annual Conference
                  on Neural Information Processing Systems 2025, NeurIPS 2025, San Diego,
                  CA, USA, December 2-7, 2025 / Mexico City, Mexico, November 30 - December
                  5, 2025},
  year         = {2025},
  url          = {http://papers.nips.cc/paper\_files/paper/2025/hash/8eb3976840e08b80dda9667562574246-Abstract-Conference.html},
  bibsource    = {dblp computer science bibliography, https://dblp.org}
}

\newpage
\appendix

\section{Additional Implementation and Training Details}
\label{appendix:additional_implementation_and_training_details}
We freeze the vision encoder, multimodal projector, and language-model backbone, and optimize only the layer-specific visual-memory adapters. To ensure a fair comparison with visual token pruning methods, we disable the DeepStack mechanism in Qwen3-VL for all methods. Unless otherwise specified, each adapter uses a bottleneck rank of $r$=128. For the default single-image setting, we train on \texttt{allenai/pixmo-ask-model-anything}~\citep{allenai_pixmo_ask_model_anything} for 2,000 steps using 8 GPUs with a global batch size of 32~(LLaVA-Mini and EPIC are trained using the same settings). For the multi-image and video setting, we construct the training set from 20,000 samples of \texttt{allenai/Molmo2-MultiImageQA}~\citep{allenai_molmo2_multiimageqa}, 44,000 samples of \texttt{lmms-lab/M4-Instruct-Data}~\citep{li2025llava_interleave}, and 64,000 samples of \texttt{Video-R1/Video-R1-data}~\citep{feng2025videor1}, and train for 4,000 steps using the same number of GPUs and global batch size. For all experiments, optimization uses AdamW with a learning rate of $5\times10^{-5}$, weight decay of 0.01, and gradient clipping at 1.0. We adopt a cosine learning-rate schedule with a 3\% warmup and a minimum learning-rate ratio of 0.1. All experiments use a fixed random seed of 44.

\section{Additional Token-Retention Results}
\label{appendix:additional_token_retention_results}
\begin{table}[htbp]
    \centering
    \caption{
    Results of visual token compression methods on
    Qwen3-VL-4B.
    Training-free baselines are evaluated under 15\% and 10\%
    visual-token retention.
    }
    \label{tab:baseline_acc_comparison_appendix}
    \resizebox{\textwidth}{!}{%
    \renewcommand{\arraystretch}{1.15}
    \begin{tabular}{lcccccccccc}
        \toprule
        \textbf{Method}
        & \textbf{MMStar}
        & \textbf{RWQA}
        & \textbf{GQA}
        & \textbf{MMB}
        & \textbf{MMB-CN}
        & \textbf{MME}
        & \textbf{POPE}
        & \textbf{SQA}
        & \textbf{VQA-v2}
        & \textbf{Avg.} \\
        \midrule

        \rowcolor{gray!15}
        \multicolumn{11}{c}{\textit{Upper Bound (100\% Retention)}} \\

        Qwen3-VL-4B
        & 64.9 & 71.2 & 61.6 & 87.5 & 87.7
        & 84.7 & 89.3 & 93.3 & 80.9 & 80.1 \\

        \midrule

        \rowcolor{gray!15}
        \multicolumn{11}{c}{\textbf{Training-Free}} \\

        \rowcolor{gray!10}
        \multicolumn{11}{c}{\textit{Retain 15\% Visual Tokens}} \\

        FastV
        & 47.5 & 50.5 & 44.1 & 76.7 & 74.8
        & 75.1 & 73.1 & 80.3 & 63.7 & 65.1 \\

        VisionZip
        & 49.7 & 58.4 & 53.4 & 80.2 & 78.4
        & 76.4 & 79.5 & 82.3 & 70.3 & 69.8 \\

        DivPrune
        & 51.0 & 62.4 & 53.3 & 83.3 & 81.9
        & 81.5 & 84.9 & 82.3 & 74.1 & 72.7 \\

        SparseVLM
        & 48.8 & 43.4 & 42.0 & 67.3 & 65.7
        & 63.4 & 73.8 & 74.3 & 65.3 & 60.4 \\

        DART
        & 48.4 & 60.5 & 51.6 & 81.8 & 81.5
        & 76.7 & 81.6 & 83.0 & 70.4 & 70.6 \\

        ZOO-Prune
        & 46.5 & 53.1 & 50.3 & 77.7 & 78.1
        & 77.6 & 76.1 & 78.1 & 69.3 & 67.4 \\

        \rowcolor{gray!10}
        \multicolumn{11}{c}{\textit{Retain 10\% Visual Tokens}} \\

        FastV
        & 43.4 & 47.8 & 43.9 & 70.7 & 68.8
        & 71.9 & 69.5 & 77.4 & 57.7 & 61.2 \\

        VisionZip
        & 46.4 & 53.9 & 50.4 & 77.4 & 75.5
        & 73.9 & 75.6 & 79.5 & 67.2 & 66.6 \\

        DivPrune
        & 48.5 & 61.4 & 51.3 & 81.1 & 81.2
        & 78.8 & 83.4 & 82.4 & 71.7 & 71.1 \\

        SparseVLM
        & 44.6 & 42.6 & 40.7 & 60.0 & 57.4
        & 61.4 & 68.7 & 72.6 & 60.3 & 56.5 \\

        DART
        & 46.0 & 55.8 & 49.4 & 77.2 & 78.4
        & 73.9 & 78.4 & 80.4 & 66.6 & 67.3 \\

        ZOO-Prune
        & 44.0 & 47.5 & 47.8 & 75.7 & 75.2
        & 75.2 & 72.7 & 77.0 & 64.3 & 64.4 \\

        \midrule

        \rowcolor{myblue}
        \textbf{\method}
        & 54.4 & 62.9 & 56.7 & 84.5 & 82.8
        & 80.1 & 87.6 & 82.3 & 78.0 & \textbf{74.4} \\

        \bottomrule
    \end{tabular}%
    }
\end{table}
To provide a more complete comparison across visual-token budgets, we additionally evaluate the training-free pruning baselines at 15\% and 10\% token retention. As shown in Table~\ref{tab:baseline_acc_comparison_appendix}, performance consistently decreases as fewer visual tokens are retained. DivPrune remains the strongest pruning baseline at both operating points, achieving average scores of 72.7 and 71.1 at 15\% and 10\% retention, respectively. In comparison, \method achieves an average score of 74.4 while preserving all visual tokens. These results provide a more complete view of the trade-off obtained by reducing the visual sequence length.

\section{Additional Result on Multi-Image and Video}
\label{appendix:additional_result_on_multi-Image_and_video}
\begin{table}[t]
    \centering
    \caption{
    Additional results on multi-image and video benchmarks.
    Training-free baselines are evaluated at 20\% visual-token retention.
    }
    \label{tab:video_multi_image_comparison_appendix}

    \resizebox{0.78\textwidth}{!}{%
    \renewcommand{\arraystretch}{1.10}
    \begin{tabular}{lcccc}
        \toprule
        \textbf{Method}
        & \textbf{MuirBench}
        & \textbf{Video-MME}
        & \textbf{MVBench}
        & \textbf{Avg.} \\
        \midrule

        \rowcolor{gray!15}
        \multicolumn{5}{c}{\textit{Upper Bound}} \\

        Qwen3-VL-4B
        & 53.5 & 52.0 & 61.6 & 55.7 \\

        \midrule

        \rowcolor{gray!10}
        \multicolumn{5}{c}{\textit{Training-Free (Retain 20\% Visual Tokens)}} \\

        FastV
        & 48.6 & 49.5 & 54.6 & 50.9 \\

        DART
        & 49.0 & 52.2 & 56.7 & 52.6 \\

        VisionZip
        & 47.9 & 50.9 & 55.0 & 51.3 \\

        SparseVLM
        & 47.0 & 49.6 & 53.9 & 50.2 \\

        DivPrune
        & 48.8 & 52.0 & 57.5 & 52.8 \\

        ZOO-Prune
        & 50.2 & 49.5 & 53.6 & 51.1 \\

        \midrule

        \rowcolor{myblue}
        Emb. Adapter (Single-Img.)
        & 40.7 & 50.5 & 57.4 & 49.5 \\

        \rowcolor{myblue}
        Emb. Adapter (Multi-Img./Video)
        & 45.5 & 51.0 & 59.5 & 52.0 \\

        \bottomrule
    \end{tabular}%
    }
\end{table}
Table~\ref{tab:video_multi_image_comparison_appendix} provides additional results on multi-image and video benchmarks, where the training-free pruning baselines retain 20\% of the visual tokens. 

\section{Additional Efficiency Results}
\label{appendix:additional efficiency_results}
\begin{table*}[htbp]
    \centering
    \caption{
    Efficiency comparison on Video-MME.
    Training-free baselines are evaluated under 20\% visual-token retention. Total Speedup measures the overall inference speedup, including both prefill and decoding.
    }
    \label{tab:prefill_speed_comparison_appendix}
    \resizebox{0.75\textwidth}{!}{%
    \renewcommand{\arraystretch}{1.15}
    \begin{tabular}{lccccc}
        \toprule
        \textbf{Method}
        & \textbf{FLOPs}
        & \textbf{Peak Mem.}
        & \textbf{Total}
        & \textbf{Prefill}
        & \textbf{Decode} \\
        & \textbf{(\%)}
        & \textbf{(GB)}
        & \textbf{Speedup}
        & \textbf{Speedup}
        & \textbf{Speedup} \\
        \midrule

        \rowcolor{gray!15}
        \multicolumn{6}{c}{\textit{Upper Bound}} \\

        Qwen3-VL-4B
        & 100.00
        & 12.94
        & 1.00
        & 1.00
        & 1.00 \\

        \midrule

        \rowcolor{gray!15}
        \multicolumn{6}{c}{\textbf{Training-Free} \textit{(Retain 20\% Visual Tokens)}} \\

        FastV
        & 33.27
        & 11.68
        & 1.18
        & 1.34
        & 1.07 \\

        DART
        & 33.00
        & 11.64
        & 1.13
        & 1.23
        & 1.06 \\

        VisionZip
        & 29.08
        & 11.05
        & 1.25
        & 1.34
        & 1.18 \\

        DivPrune
        & 29.05
        & 11.06
        & 1.25
        & 1.34
        & 1.18 \\

        ZOO-Prune
        & 36.36
        & 12.19
        & 1.16
        & 1.15
        & 1.18 \\

        SparseVLM
        & 33.37
        & 11.98
        & 1.16
        & 1.28
        & 1.07 \\

        \midrule

        \rowcolor{myblue}
        Emb. Adapter
        & 17.90
        & 12.97
        & 1.30
        & 1.50
        & 1.13 \\

        \bottomrule
    \end{tabular}%
    }
\end{table*}
Table~\ref{tab:prefill_speed_comparison_appendix} reports the inference efficiency of different visual token pruning methods under 20\% token retention on Video-MME. The pruning baselines require 29.05–36.36\% of the FLOPs of the uncompressed model and achieve prefill speedups ranging from $1.15\times$ to $1.34\times$. Their total speedups range from $1.13\times$ to $1.25\times$, while decode speedups remain between $1.06\times$ and $1.18\times$. \method uses 17.90\% of the FLOPs and achieves $1.30\times$ total and $1.50\times$ prefill speedups while preserving all visual tokens. These results provide an additional efficiency comparison under a mild pruning regime.

\section{Training Objective Ablation}
\label{appendix:training_objective_ablation}
\begin{table}[htbp]
    \centering
    \caption{
    Ablation study on different training objectives for the embedding adapter.
    All variants only optimize the adapter parameters. Emb. Adapter (Init). denotes the initialization setting, where the same visual representation is shared across all Transformer layers without learned layer-specific adaptation, serving as the lower-bound baseline.
    }
    \label{tab:training_objective_comparison}
    \resizebox{\textwidth}{!}{%
    \renewcommand{\arraystretch}{1.15}
    \begin{tabular}{lccccccccccc}
        \toprule
        \textbf{Method}
        & \textbf{MMStar}
        & \textbf{RWQA}
        & \textbf{GQA}
        & \textbf{MMB}
        & \textbf{MMB-CN}
        & \textbf{MME}
        & \textbf{POPE}
        & \textbf{SQA}
        & \textbf{VQA-v2}
        & \textbf{Avg.}
        & \textbf{Time} \\
        \midrule

        Emb. Adapter (Init).
        & 41.2
        & 40.8
        & 43.4
        & 77.4
        & 74.6
        & 66.4
        & 65.5
        & 78.8
        & 58.4
        & 60.7
        & - \\

        Emb. Adapter + SFT
        & 50.5
        & 59.4
        & 28.6
        & 83.6
        & 80.0
        & 77.5
        & 85.5
        & 79.9
        & 75.1
        & 68.9
        & \textbf{23m51s} \\

        Emb. Adapter + OPD
        & 53.6
        & \textbf{63.9}
        & 55.8
        & 83.4
        & \textbf{83.1}
        & 79.5
        & 86.7
        & 81.9
        & 77.8
        & 74.0
        & 4h38m08s \\

        \rowcolor{myblue}
        \textbf{Emb. Adapter + Supervised-KD}
        & \textbf{54.4}
        & 62.9
        & \textbf{56.7}
        & \textbf{84.5}
        & 82.8
        & \textbf{80.1}
        & \textbf{87.6}
        & \textbf{82.3}
        & \textbf{78.0}
        & \textbf{74.4}
        & 34m53s \\

        \bottomrule
    \end{tabular}%
    }
\end{table}
We compare three objectives for training the embedding adapter while keeping all backbone parameters frozen: supervised fine-tuning (SFT), on-policy distillation (OPD)~\citep{agarwal2024policy}, and our Supervised-KD objective. As shown in Table~\ref{tab:training_objective_comparison}, SFT achieves an average score of 68.9, substantially below both distillation-based objectives. OPD improves the average score to 74.0 but requires 4h38m of training due to autoregressive student rollouts. Supervised-KD achieves the highest average score of 74.4 while requiring only 34m53s, reducing the training time by approximately $8\times$ relative to OPD. These results show that teacher-forced distribution matching retains the effectiveness of distillation without the rollout overhead of on-policy training.

\section{Additional Cross-Backbone Results}
\label{appendix:additional_cross_backbone_results}
\begin{table*}[htbp]
    \centering
    \caption{
    Results of different visual token compression methods across
    various vision-language models.
    \textit{Vanilla} denotes the uncompressed upper bound.
    DART and DivPrune retain 20\% of the visual tokens.
    }
    \label{tab:different_models_acc_comparison_appendix}

    \scriptsize
    \setlength{\tabcolsep}{2.0pt}
    \renewcommand{\arraystretch}{1.05}

    \begin{tabular}{llcccccccccc}
        \toprule
        \textbf{Model}
        & \textbf{Method}
        & \textbf{MMStar}
        & \textbf{RWQA}
        & \textbf{GQA}
        & \textbf{MMB}
        & \textbf{MMB-CN}
        & \textbf{MME}
        & \textbf{POPE}
        & \textbf{SQA}
        & \textbf{VQA-v2}
        & \textbf{Avg.} \\
        \midrule

        \multirow{4}{*}{\textbf{LLaVA-1.5-7B}}
        & Vanilla
        & 37.0 & 54.9 & 61.2 & 72.5 & 67.9
        & 71.7 & 83.4 & 65.6 & 75.9 & 65.6 \\

        & DART
        & 30.6 & 52.3 & 57.8 & 70.8 & 66.3
        & 76.6 & 81.4 & 69.4 & 71.1 & 64.0 \\

        & DivPrune
        & 38.1 & 51.9 & 58.5 & 70.4 & 66.5
        & 70.3 & 82.1 & 65.6 & 74.3 & \textbf{64.2} \\

        \rowcolor{myblue}
        & \textbf{\method}
        & 35.4 & 51.4 & 53.8 & 69.9 & 63.7
        & 68.7 & 79.2 & 65.4 & 69.5 & 61.9 \\

        \midrule

        \multirow{4}{*}{\textbf{LLaVA-1.5-13B}}
        & Vanilla
        & 38.7 & 57.7 & 63.0 & 74.7 & 70.3
        & 66.6 & 83.8 & 71.5 & 78.3 & 67.2 \\

        & DART
        & 33.6 & 55.4 & 57.6 & 76.0 & 67.9
        & 75.6 & 84.9 & 73.6 & 72.1 & \textbf{66.3} \\

        & DivPrune
        & 36.8 & 52.3 & 60.8 & 73.8 & 69.6
        & 69.1 & 81.2 & 72.2 & 75.8 & 65.7 \\

        \rowcolor{myblue}
        & \textbf{\method}
        & 36.5 & 50.1 & 56.3 & 72.5 & 65.9
        & 64.2 & 76.8 & 70.7 & 73.1 & 62.9 \\

        \midrule

        \multirow{4}{*}{\textbf{LLaVA-1.6-Mistral-7B}}
        & Vanilla
        & 40.4 & 47.3 & 54.3 & 71.9 & 65.0
        & 68.4 & 85.4 & 68.7 & 74.4 & 64.0 \\

        & DART
        & 40.0 & 48.1 & 53.2 & 72.0 & 64.2
        & 69.8 & 84.4 & 69.8 & 73.6 & 63.9 \\

        & DivPrune
        & 39.4 & 47.1 & 56.8 & 72.1 & 63.8
        & 68.2 & 83.4 & 70.5 & 75.6 & \textbf{64.1} \\

        \rowcolor{myblue}
        & \textbf{\method}
        & 39.2 & 43.9 & 46.8 & 70.5 & 61.2
        & 64.8 & 83.7 & 68.3 & 69.5 & 60.9 \\

        \midrule

        \multirow{4}{*}{\textbf{Qwen3-VL-8B}}
        & Vanilla
        & 68.5 & 69.4 & 61.9 & 89.2 & 88.7
        & 88.6 & 86.8 & 95.1 & 80.6 & 81.0 \\

        & DART
        & 51.3 & 63.5 & 52.8 & 81.9 & 81.6
        & 80.4 & 82.6 & 88.0 & 71.8 & 72.7 \\

        & DivPrune
        & 54.4 & 64.1 & 56.2 & 83.4 & 84.4
        & 84.2 & 85.9 & 86.0 & 76.8 & \textbf{75.0} \\

        \rowcolor{myblue}
        & \textbf{\method}
        & 54.9 & 61.8 & 58.0 & 83.5 & 82.6
        & 82.0 & 84.6 & 81.5 & 75.4 & 73.8 \\

        \midrule

        \multirow{4}{*}{\textbf{Qwen3-VL-30B-A3B}}
        & Vanilla
        & 70.3 & 72.3 & 65.2 & 89.4 & 90.8
        & 89.4 & 87.9 & 96.3 & 82.6 & 82.7 \\

        & DART
        & 58.3 & 64.8 & 59.6 & 85.9 & 85.4
        & 86.1 & 83.3 & 89.5 & 75.9 & 76.5 \\

        & DivPrune
        & 59.6 & 68.0 & 62.8 & 86.7 & 87.7
        & 87.5 & 88.0 & 90.3 & 78.7 & \textbf{78.8} \\

        \rowcolor{myblue}
        & \textbf{\method}
        & 62.4 & 65.1 & 62.4 & 86.5 & 86.8
        & 87.0 & 87.8 & 88.6 & 80.7 & 78.6 \\

        \midrule

        \multirow{4}{*}{\textbf{Qwen3.5-4B}}
        & Vanilla
        & 48.9 & 67.2 & 63.4 & 84.3 & 77.2
        & 84.0 & 88.7 & 78.4 & 80.0 & 74.7 \\

        & DART
        & 37.3 & 56.5 & 54.0 & 79.5 & 73.5
        & 79.7 & 79.2 & 73.4 & 71.6 & 67.2 \\

        & DivPrune
        & 42.3 & 63.3 & 57.7 & 81.1 & 73.3
        & 81.9 & 87.3 & 74.1 & 75.9 & 70.8 \\

        \rowcolor{myblue}
        & \textbf{\method}
        & 40.1 & 65.8 & 60.5 & 81.3 & 75.1
        & 79.7 & 89.0 & 74.8 & 77.7 & \textbf{71.6} \\

        \bottomrule
    \end{tabular}
\end{table*}

Table~\ref{tab:different_models_acc_comparison_appendix} reports cross-backbone results with DART and DivPrune retaining 20\% of the visual tokens. Across the 6 evaluated backbones, \method remains close to the strongest pruning baseline on the Qwen models, achieving 73.8 on Qwen3-VL-8B, 78.6 on Qwen3-VL-30B-A3B, and 71.6 on Qwen3.5-4B, compared with 75.0, 78.8, and 70.8 for DivPrune, respectively. On the LLaVA family, \method obtains average scores of 61.9, 62.9, and 60.9 on LLaVA-1.5-7B, LLaVA-1.5-13B, and LLaVA-v1.6-Mistral-7B. Table~\ref{tab:different_models_acc_comparison_appendix_2} shows the results under 5\% Token Retention. \method achieves 62.9 on LLaVA-1.5-13B and 73.8 on Qwen3-VL-8B, improving over DivPrune by 6.9 and 8.4 points, respectively. On LLaVA-v1.6-Mistral-7B, \method reaches 60.9, which is comparable to DivPrune at 61.5 and substantially higher than DART at 50.4.

\begin{table*}[htbp]
    \centering
    \caption{
    Results of different visual token compression methods across
    various vision-language models. \textit{Vanilla} denotes the uncompressed
    upper bound. DART and DivPrune retain 5\% of the visual tokens.
    }
    \label{tab:different_models_acc_comparison_appendix_2}

    \scriptsize
    \setlength{\tabcolsep}{2.0pt}
    \renewcommand{\arraystretch}{1.05}

    \begin{tabular}{llcccccccccc}
        \toprule
        \textbf{Model}
        & \textbf{Method}
        & \textbf{MMStar}
        & \textbf{RWQA}
        & \textbf{GQA}
        & \textbf{MMB}
        & \textbf{MMB-CN}
        & \textbf{MME}
        & \textbf{POPE}
        & \textbf{SQA}
        & \textbf{VQA-v2}
        & \textbf{Avg.} \\
        \midrule

        \multirow{4}{*}{\textbf{LLaVA-1.5-13B}}
        & Vanilla
        & 38.7 & 57.7 & 63.0 & 74.7 & 70.3
        & 66.6 & 83.8 & 71.5 & 78.3 & 67.2 \\

        & DART
        & 32.7 & 48.1 & 50.0 & 67.7 & 58.2
        & 58.4 & 70.7 & 70.9 & 63.0 & 57.7 \\

        & DivPrune
        & 35.9 & 51.2 & 44.1 & 69.5 & 60.9
        & 52.8 & 62.9 & 68.3 & 58.5 & 56.0 \\

        \rowcolor{myblue}
        & \textbf{\method}
        & 36.5 & 50.1 & 56.3 & 72.5 & 65.9
        & 64.2 & 76.8 & 70.7 & 73.1 & \textbf{62.9} \\

        \midrule

        \multirow{4}{*}{\textbf{LLaVA-1.6-Mistral-7B}}
        & Vanilla
        & 40.4 & 47.3 & 54.3 & 71.9 & 65.0
        & 68.4 & 85.4 & 68.7 & 74.4 & 64.0 \\

        & DART
        & 29.6 & 42.5 & 39.1 & 51.0 & 46.1
        & 58.6 & 67.0 & 67.5 & 52.3 & 50.4 \\

        & DivPrune
        & 36.8 & 42.9 & 54.7 & 69.4 & 61.3
        & 63.7 & 84.6 & 69.6 & 70.5 & \textbf{61.5} \\

        \rowcolor{myblue}
        & \textbf{\method}
        & 39.2 & 43.9 & 46.8 & 70.5 & 61.2
        & 64.8 & 83.7 & 68.3 & 69.5 & 60.9 \\

        \midrule

        \multirow{4}{*}{\textbf{Qwen3-VL-8B}}
        & Vanilla
        & 68.5 & 69.4 & 61.9 & 89.2 & 88.7
        & 88.6 & 86.8 & 95.1 & 80.6 & 81.0 \\

        & DART
        & 40.6 & 53.3 & 43.5 & 70.7 & 71.7
        & 74.5 & 74.4 & 79.3 & 61.5 & 63.3 \\

        & DivPrune
        & 41.0 & 55.3 & 44.9 & 73.4 & 74.6
        & 76.8 & 76.8 & 78.4 & 67.2 & 65.4 \\

        \rowcolor{myblue}
        & \textbf{\method}
        & 54.9 & 61.8 & 58.0 & 83.5 & 82.6
        & 82.0 & 84.6 & 81.5 & 75.4 & \textbf{73.8} \\

        \bottomrule
    \end{tabular}
\end{table*}

\section{Adapter Rank Ablation}
\begin{figure}[htbp]
    \centering
    \begin{subfigure}[t]{0.49\linewidth}
        \centering
        \includegraphics[width=\linewidth]{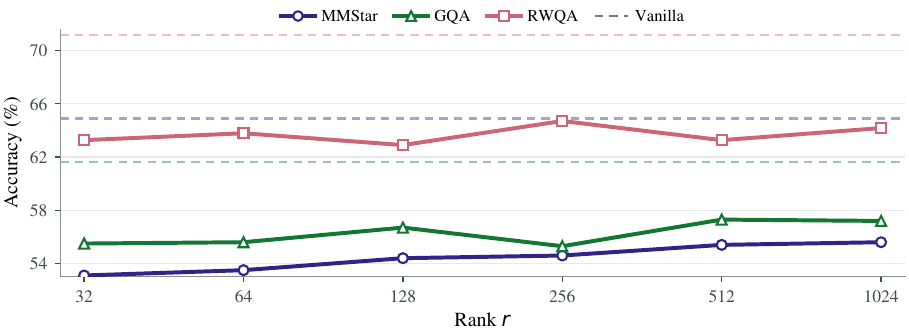}
        \caption{Results on MMStar, GQA, and RWQA.}
        \label{fig:rank_low}
    \end{subfigure}
    \hfill
    \begin{subfigure}[t]{0.49\linewidth}
        \centering
        \includegraphics[width=\linewidth]{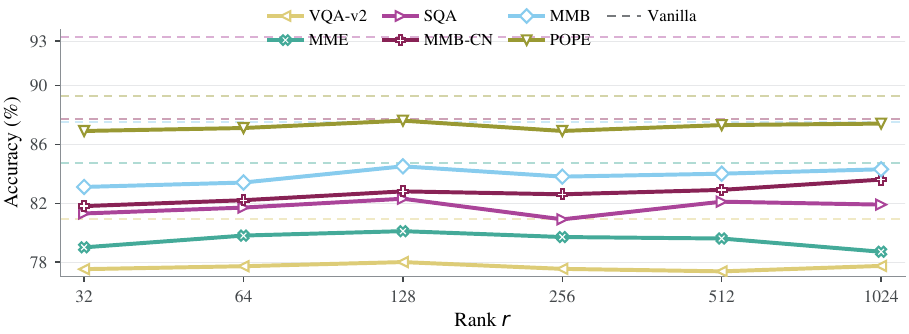}
        \caption{Results on VQA-v2, MME, SQA, MMB-CN, MMB, and POPE.}
        \label{fig:rank_high}
    \end{subfigure}

    \caption{
    Performance of Embedding Adapter under different adapter ranks $r$. Dashed horizontal lines denote the corresponding vanilla model performance.
    }
    \label{fig:rank_comparison}
\end{figure}

We further study the effect of the bottleneck rank $r$ of the visual-memory adapter. As shown in Figure~\ref{fig:rank_comparison}, increasing the adapter rank generally leads to
moderate improvements in downstream accuracy, suggesting that a larger bottleneck provides additional capacity for modeling layer-specific visual-state corrections. 

\section{Hidden-State Compression}
\label{appendix:hidden_state_compression}

To construct the low-rank subspace, we collect visual hidden states from 1,024 images in \texttt{allenai/pixmo-ask-model-anything} and fit a separate uncentered PCA basis for each Transformer layer. The basis is shared across all visual tokens and images at the same layer. Within a given layer, the same basis is shared across all visual-token positions and all input samples. At inference time, each visual hidden state $h \in \mathbb{R}^{D}$ at layer $l$ is projected onto the leading $r$ principal directions and then reconstructed back to the original hidden dimension:
\begin{equation}
      \tilde{h} = P_{l,r}P_{l,r}^{\top}h,
\end{equation}
where $P_{l,r}\in\mathbb{R}^{D\times r}$ contains the leading $r$ principal directions for layer $l$. The reconstructed state $\tilde{h}$ then replaces the original visual hidden state for subsequent computation. This intervention preserves the number and positions of visual tokens and changes only the dimensional subspace in which their hidden representations are allowed to vary. For different values of $r$, we use nested prefixes of the same PCA basis, and the fitted bases are shared across all evaluation benchmarks.

\section{Adapter Compatibility}
\label{appendix:adapter_combine_with_token_pruning_methods}
\begin{table*}[htbp]
    \centering
    \caption{
    Combining visual-memory adaptation with visual-token pruning on Qwen3-VL-4B.
    Starting from the embedding adapter, we additionally apply DART or DivPrune
    at different visual-token retention ratios.
    }
    \label{tab:adapter_pruning_compatibility}

    \setlength{\tabcolsep}{4.2pt}
    \renewcommand{\arraystretch}{1.10}

    \begin{adjustbox}{width=\textwidth}
    \begin{tabular}{
        l
        c
        c c c c c c c c c
        >{\columncolor{gray!8}}c
    }
        \toprule

        \textbf{Method}
        & \textbf{Retention}
        & \textbf{MMStar}
        & \textbf{RWQA}
        & \textbf{GQA}
        & \textbf{MMB}
        & \textbf{MMB-CN}
        & \textbf{MME}
        & \textbf{POPE}
        & \textbf{SQA}
        & \textbf{VQA-v2}
        & \textbf{Avg.} \\

        \midrule

        Vanilla
        &
        & 64.9 & 71.2 & 61.6 & 87.5 & 87.7
        & 84.7 & 89.3 & 93.3 & 80.9
        & 80.1 \\

        \midrule

        \textbf{\method (Emb. Adapter)}
        &
        & 54.4 & 62.9 & 56.7 & 84.5 & 82.8
        & 80.1 & 87.6 & 82.3 & 78.0
        & 74.4 \\

        \hspace{1.0em}$+$ DART
        & \multirow{2}{*}{50\%}
        & 49.7 & 62.1 & 58.4 & 82.6 & 80.9
        & 81.8 & 88.0 & 81.0 & 74.3
        & 73.2 \\

        \hspace{1.0em}$+$ DivPrune
        &
        & 48.8 & 61.4 & 58.4 & 83.0 & 80.7
        & 82.7 & 87.1 & 81.3 & 74.5
        & 73.1 \\

        \midrule

        \hspace{1.0em}$+$ DART
        & \multirow{2}{*}{20\%}
        & 46.4 & 58.4 & 54.1 & 79.5 & 78.9
        & 79.1 & 83.9 & 79.6 & 68.5
        & 69.8 \\

        \hspace{1.0em}$+$ DivPrune
        &
        & 45.1 & 57.4 & 55.2 & 79.5 & 78.5
        & 82.4 & 84.8 & 79.4 & 70.2
        & 70.3 \\

        \midrule

        \hspace{1.0em}$+$ DART
        & \multirow{2}{*}{5\%}
        & 39.5 & 50.2 & 45.1 & 69.0 & 70.9
        & 72.8 & 70.9 & 76.4 & 58.0
        & 61.4 \\

        \hspace{1.0em}$+$ DivPrune
        &
        & 36.9 & 53.2 & 45.2 & 67.9 & 67.6
        & 75.2 & 69.4 & 75.8 & 58.1
        & 61.0 \\

        \bottomrule
    \end{tabular}
    \end{adjustbox}

\end{table*}
Our method reduces visual computation along the representation dimension,
whereas token-pruning methods reduce the visual sequence length. We therefore further investigate whether these two forms of compression can be applied jointly. Specifically, starting from the embedding-adapter variant of \method, we additionally apply DART or DivPrune at different visual-token retention ratios. The results are reported in
Table~\ref{tab:adapter_pruning_compatibility}. The embedding adapter alone achieves an average score of 74.4. When combined with moderate token pruning at 50\% retention, performance remains largely
preserved: DART and DivPrune achieve average scores of 73.2 and 73.1,
respectively, corresponding to only a 1.2--1.3 point decrease from the
standalone adapter. More aggressive pruning leads to the expected degradation, with average scores of 69.8/70.3 at 20\% retention and 61.4/61.0 at 5\% retention for DART/DivPrune. These results provide empirical evidence that reducing layer-wise visual-state computation and reducing visual sequence length target distinct sources of redundancy and can be used in a complementary manner.

\section{Layer-wise Adapter Skipping}
\label{appendix:adapter_layer_skipping}
\begin{table*}[htbp]
    \centering
    \caption{
    Effect of layer-wise adapter skipping on Qwen3-VL-4B.
    We selectively skip the embedding adapters in early and late Transformer
    layers while keeping the remaining configuration unchanged.
    }
    \label{tab:adapter_layer_ablation}

    \setlength{\tabcolsep}{4.0pt}
    \renewcommand{\arraystretch}{1.08}

    \begin{adjustbox}{width=0.90\textwidth}
    \begin{tabular}{
        l
        c c c c c c c c c
        >{\columncolor{gray!8}}c
    }
        \toprule

        \textbf{Configuration}
        & \textbf{MMStar}
        & \textbf{RWQA}
        & \textbf{GQA}
        & \textbf{MMB}
        & \textbf{MMB-CN}
        & \textbf{MME}
        & \textbf{POPE}
        & \textbf{SQA}
        & \textbf{VQA-v2}
        & \textbf{Avg.} \\

        \midrule

        \textbf{Emb. Adapter}
        & 54.4
        & 62.9
        & 56.7
        & 84.5
        & 82.8
        & 80.1
        & 87.6
        & 82.3
        & 78.0
        & 74.4 \\

        \hspace{1.2em} w/o First 5 \& Last 10
        & 49.6
        & 62.5
        & 56.6
        & 83.4
        & 81.1
        & 82.1
        & 89.0
        & 81.1
        & 74.3
        & 73.3 \\

        \hspace{1.2em} w/o First 10 \& Last 10
        & 46.7
        & 61.2
        & 55.4
        & 82.8
        & 78.9
        & 80.8
        & 88.3
        & 80.5
        & 72.4
        & 71.9 \\

        \bottomrule
    \end{tabular}
    \end{adjustbox}
\end{table*}
Motivated by the non-uniform layer-wise dependence on visual information observed in Table~\ref{tab:intervention_layer_low_rank_acc}, we further investigate whether the visual-memory adapters need to be applied at every Transformer layer. We selectively skip the embedding adapters in the early and late layers while keeping the remaining configuration unchanged. As shown in Table~\ref{tab:adapter_layer_ablation}, removing the adapters from the first 5 and last 10 layers reduces the average score only from 74.4 to 73.3, while skipping the first 10 and last 10 layers yields an average score
of 71.9. Meanwhile, Table~\ref{tab:adapter_skipping_efficiency} shows that layer-wise skipping further reduces the computational overhead of \method: the FLOPs decrease from 17.90\% to 15.64\% and 14.89\% of the vanilla model, while the total inference
speedup improves from $1.30\times$ to $1.35\times$ and $1.37\times$, respectively. These results are consistent with the depth-wise intervention analysis and suggest that visual-memory adaptation need not be allocated uniformly across layers. In particular, selectively skipping adapters in less sensitive early and late layers provides an additional accuracy--efficiency trade-off for \method.
\begin{table*}[htbp]
    \centering
    \caption{
    Efficiency of layer-wise adapter skipping on Qwen3-VL-4B.
    Total Speedup measures the overall inference speedup, including both
    prefill and decoding.
    }
    \label{tab:adapter_skipping_efficiency}

    \setlength{\tabcolsep}{6pt}
    \renewcommand{\arraystretch}{1.08}

    \begin{adjustbox}{width=0.72\textwidth}
    \begin{tabular}{
        l
        >{\columncolor{gray!8}}c
        c
        c
        c
        c
    }
        \toprule

        \textbf{Configuration}
        & \textbf{FLOPs}
        & \textbf{Peak Mem.}
        & \textbf{Total}
        & \textbf{Prefill}
        & \textbf{Decode} \\

        & \textbf{(\%)}
        & \textbf{(GB)}
        & \textbf{Speedup}
        & \textbf{Speedup}
        & \textbf{Speedup} \\

        \midrule

        Vanilla
        & 100.00
        & 12.94
        & 1.00
        & 1.00
        & 1.00 \\

        \midrule

        \textbf{Emb. Adapter}
        & 17.90
        & 12.97
        & 1.30
        & 1.50
        & 1.13 \\

        \hspace{1.2em} w/o First 5 \& Last 10
        & 15.64
        & 11.86
        & 1.35
        & 1.60
        & 1.16 \\

        \hspace{1.2em} w/o First 10 \& Last 10
        & 14.89
        & 11.49
        & 1.37
        & 1.64
        & 1.17 \\

        \bottomrule
    \end{tabular}
    \end{adjustbox}

\end{table*}

\section{Linear Attention Analysis}
\label{appendix:linear_attention_analysis}
For hybrid backbones containing Gated DeltaNet layers, such as Qwen3.5-4B, \method independently constructs the visual memory at each layer. Under the embedding-adapter variant, the predicted representation of the $i$-th visual token at layer $l$ is
\begin{equation}
M_{l,i}
=
E_i
+
U_{l}
\operatorname{SiLU}
\left(
D_{l}E_i
\right).
\end{equation}
Thus, the visual representation at each layer is constructed directly from the initial visual embedding $E_i$, rather than propagated from the visual state of the preceding layer. At layer $l$, we replace the visual positions with the predicted visual memory while keeping the textual hidden states unchanged:
\begin{equation}
X_{l,t}
=
\begin{cases}
M_{l,i}, & t \text{ corresponds to the } i\text{-th visual token},\\
H_{l,t}, & t \text{ corresponds to a textual token}.
\end{cases}
\end{equation}
The combined sequence $X_l$ is then processed by the original frozen RMSNorm, linear projections, causal convolution, and gating modules of Gated DeltaNet to obtain
$q_t$, $k_t$, $v_t$, $\alpha_t$, and $\beta_t$.

The predicted visual tokens still participate in the native recurrent-state update. Omitting the layer and head indices for clarity, the recurrence is
\begin{equation}
\bar{S}_t
=
\alpha_t S_{t-1},
\end{equation}
followed by
\begin{equation}
S_t
=
\bar{S}_t
+
\beta_t k_t
\left(
v_t-\bar{S}_t^{\top}k_t
\right)^{\top}.
\end{equation}
The corresponding readout is
\begin{equation}
o_t
=
\frac{q_t^{\top}S_t}{\sqrt{d_k}}.
\end{equation}
Therefore, the visual representations predicted by the adapter are still written into the recurrent state through the original Gated DeltaNet update, and subsequent textual tokens retrieve the resulting visual information through their own queries.

\begin{figure}[t]
    \centering
    \includegraphics[width=0.6\linewidth]{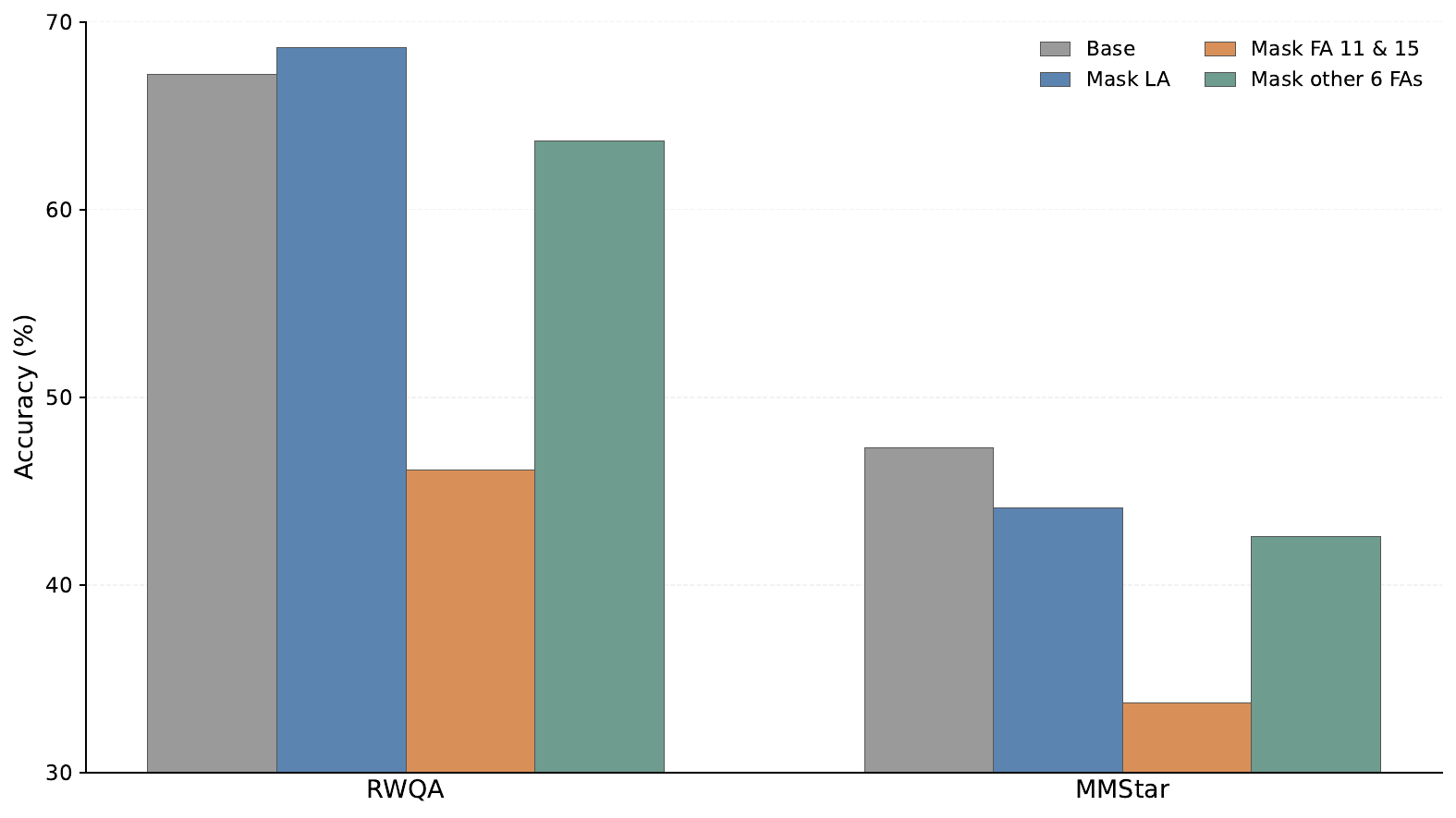}
    \caption{
    \textbf{Visual write and read interventions in the hybrid-attention backbone.}
    Mask LA removes the visual write to the recurrent state in all Linear Attention (LA) layers by restoring the post-visual recurrent state to its pre-visual value.
    For Full Attention (FA) layers, we block visual read by preventing textual queries from attending to visual keys and values. We compare blocking FA layers 11 and 15 against blocking the other six FA layers (3, 7, 19, 23, 27, and 31).
    }
    \label{fig:la_fa_intervention}
\end{figure}

To better understand how visual information is utilized in hybrid-attention MLLMs, we separately intervene on the two mechanisms through which visual information affects subsequent textual computation: writing visual information into the recurrent state of Linear Attention (LA) layers, and reading visual key--value states in Full Attention (FA) layers.

\paragraph{Blocking visual write in Linear Attention.}
For each LA layer, let $S_{\mathrm{in}}$ denote the recurrent state immediately before the first visual token, and let $S_{\mathrm{out}}$ denote the state after all visual tokens have been processed normally.
Under our \emph{rank-0} intervention, we restore the recurrent state at the visual boundary as
\begin{equation}
    S_{\mathrm{out}} \leftarrow S_{\mathrm{in}}.
\end{equation}
Thus, the visual segment is still fully processed, but its net contribution to the recurrent memory is removed before subsequent textual tokens are processed. All FA layers remain unmodified and can still access visual keys and values.
We apply this intervention to all 24 LA layers.

\paragraph{Blocking visual read in Full Attention.}
For FA layers, we instead prevent textual queries from reading visual keys and values and restricted to textual keys and values only. All LA layers operate normally.
\begin{equation}
    o_t =
    \mathrm{Attn}
    \left(
        q_t,
        K_{\mathrm{text}},
        V_{\mathrm{text}}
    \right).
\end{equation}
We evaluate two FA intervention settings.
The first blocks visual read in layers 11 and 15, while the second blocks visual read in layers 3, 7, 19, 23, 27, and 31. All layer indices are zero-based. 
As shown in Figure~\ref{fig:la_fa_intervention}, it reveals a strongly non-uniform dependence on visual information across the hybrid-attention layers. Canceling visual writes to the recurrent states of all 24 LA layers has only a limited effect, In contrast, blocking visual reads in only FA layers 11 and 15 causes a substantial degradation on RWQA and MMStar. Blocking the other six FA layers (3, 7, 19, 23, 27, and 31) results in a considerably smaller drop. These results indicate that useful visual information is highly concentrated in a small subset of FA layers rather than being uniformly utilized throughout the network. This non-uniformity provides additional flexibility for \method: visual-memory computation can be selectively retained in sensitive layers while being skipped in less important layers to further improve efficiency.


\end{document}